\documentclass[journal]{IEEEtran}
\ifCLASSINFOpdf
  \usepackage[pdftex]{graphicx}
\else
  \usepackage[dvips]{graphicx}
\fi
\usepackage{amsmath}
\usepackage{amssymb}
\usepackage{amsfonts}
\usepackage{multirow}
\usepackage{array}
\usepackage{url}

\begin{document}
%
\title{Semantic Signatures for Large-scale Visual Localization}
%
%
%

\author{Li~Weng,~\IEEEmembership{}
        Val\'erie Gouet-Brunet,~\IEEEmembership{}        
        and~Bahman Soheilian~\IEEEmembership{}
\thanks{L. Weng is with the School of Automation (Artificial Intelligence), Hangzhou Dianzi University, Hangzhou, 310018 China. V. Gouet-Brunet and B. Soheilian are with Univ. Gustave Eiffel, ENSG, IGN, Saint-Mande, 94160 France. This research was supported by Zhejiang Provincial Natural Science Foundation of China under Grant No. LY19F030022, and the European project KET ENIAC Things2Do under ENIAC JU grant agreement No. 621221.}
}

\maketitle

\begin{abstract}
Visual localization is a useful alternative to standard localization techniques. It works by utilizing cameras. In a typical scenario, features are extracted from captured images and compared with geo-referenced databases. Location information is then inferred from the matching results. Conventional schemes mainly use low-level visual features. These approaches offer good accuracy but suffer from scalability issues. In order to assist localization in large urban areas, this work explores a different path by utilizing high-level semantic information. It is found that object information in a street view can facilitate localization. A novel descriptor scheme called ``semantic signature'' is proposed to summarize this information. A semantic signature consists of type and angle information of visible objects at a spatial location. Several metrics and protocols are proposed for signature comparison and retrieval. They illustrate different trade-offs between accuracy and complexity. Extensive simulation results confirm the potential of the proposed scheme in large-scale applications. This paper is an extended version of a conference paper in CBMI'18. A more efficient retrieval protocol is presented with additional experiment results.
\end{abstract}


%
\IEEEpeerreviewmaketitle

\section{Introduction}
\begin{figure*}
\centering
\begin{tabular}{cc}
\includegraphics[scale=0.3]{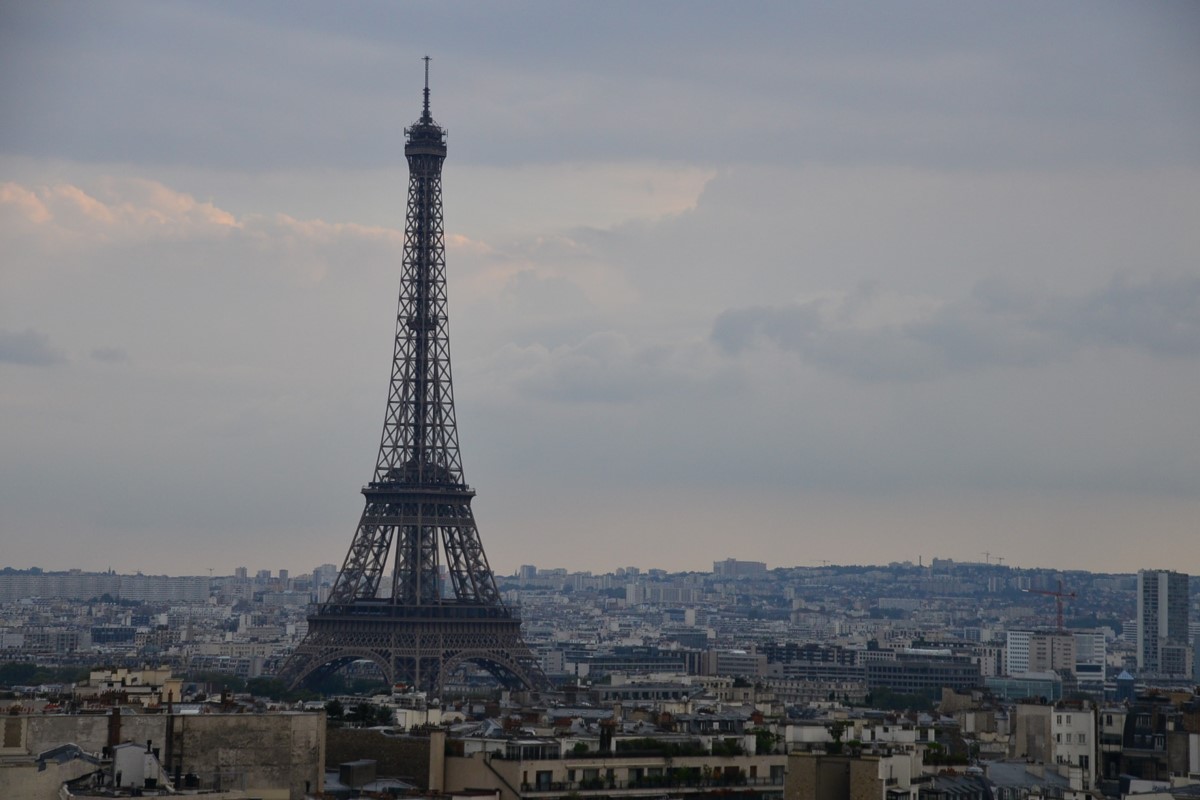} & \includegraphics[scale=0.16]{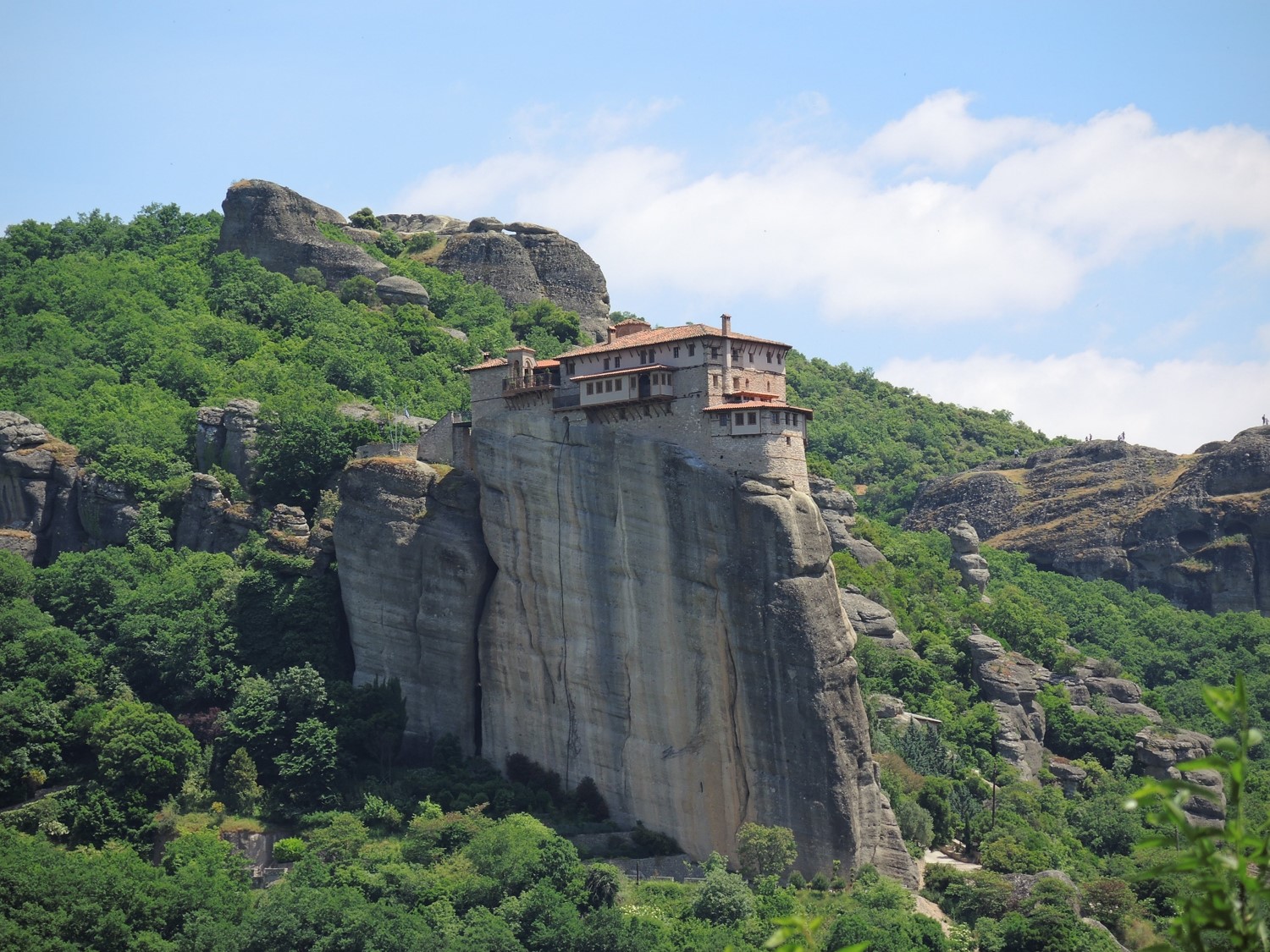}\\
(a) & (b)
\end{tabular}
\caption{Example query images for visual localization.}
\label{fig_query_image}
\end{figure*}

Visual localization~\cite{Lowry2016,Piasco2018} represents a range of applications where location information is derived from images. As an alternative to conventional positioning solutions, visual localization finds potential applications in automatic navigation~\cite{Lim2012} and location-related multimedia service~\cite{Snavely2006,Crandall2009}, such as landmark recognition and augmented reality (AR). For example, if a landmark recognition system is given the photos in Fig.~\ref{fig_query_image}, it might be able to return the landmark name, the city name, or the coordinate. In general, the problem of visual localization is to infer where an image is acquired by matching it with a geo-referenced database. It is typically modelled as an image feature retrieval scenario, and solved by exact or approximate nearest neighbour search. More specifically, features are extracted from a query image and compared with features in a database; the location is inferred from the best matches.
Depending on the required accuracy, a visual localization algorithm is designed for one of following tasks:
\begin{itemize}
\item Place recognition (coarse localization);
\item Camera pose estimation (precise localization).
\end{itemize}
The former estimates the zone where the image was acquired, in the form of a spatial area, a semantic label, etc.~\cite{Lowry2016}; the latter estimates the camera pose up to six degrees of freedom (6-DOF), including three parameters of translation (x,y,z) and three parameters of rotation (pitch, roll, yaw)~\cite{Qu2015,Brachmann2018}.
Conventional schemes typically accomplish these tasks using low-level hand-crafted visual features, such as bag of SIFT features~\cite{Lowe2004}, and more recently learned features \cite{Piasco2018}.
Related research mainly focuses on accuracy and efficiency in challenging conditions (e.g. season change, night/day, long-term datasets). Various efforts have been devoted to database indexing and query strategies~\cite{Schindler2007,Li2010,Sattler2011,Li2012,Lowry2016,Piasco2018}.
They offer good accuracy but suffer from scalability issues due to large amounts of data.

In this work, a novel approach is pursued to complement conventional visual localization. Instead of low-level visual features, we exploit high-level semantic features, which are related to what we see in the environment. For example, in dense urban areas, one can typically see buildings, cars, trees, etc. It is found that such information can facilitate localization too. 
Compared with conventional visual features, semantic features have several advantages. First, they can be encoded in a compact way and require much less storage. Second, they can be efficiently obtained from geographic information systems, such as OpenStreetMap.
Nevertheless, how to represent and utilize semantic features in localization remains an open problem. This paper summarizes our effort to address the encoding and comparison of such information.

In particular, we focus on ``semantic objects'' which are static and widely available in urban areas. We assume that such objects can be detected from street-view images by object detection algorithms such as~\cite{Lin2017,Redmon2018}. Once they are detected, localization can be achieved by finding locations with similar objects from a database. In order to represent the distribution of semantic objects at a location, we propose a descriptor called ``semantic signature''. It is a compact string that consists of type and angle information of street objects. We then model our localization problem as string matching and solve it with a retrieval framework.
Compared with conventional approaches, the proposed scheme has a few advantages, including small database size, large coverage area, and fast retrieval.

This paper is an extended version of~\cite{Weng2018}, where the semantic signature is originally proposed to summarize semantic objects. The contribution of~\cite{Weng2018} also includes suitable metrics and a ``metric fusion'' protocol for signature matching and retrieval, and a simulation framework for performance evaluation. In this paper, a more efficient retrieval protocol called ``2-stage metric fusion'' is presented, as well as additional experiment results. Although there is still a gap from practical deployment, promising results from extensive simulation indicate the potential of our approach in large-scale applications.

The rest of the paper is organized as follows: Section~\ref{background} introduces the motivation of our approach and the role of semantic information in a big picture of visual localization; Section~\ref{sect_related_work} is a brief literature review; Section~\ref{sect_proposed} describes the proposed method; Section~\ref{sec:experiments} shows experiment results and analysis; Section~\ref{sect_conclusion} concludes the work with a discussion.

\section{Background}
\label{background}
Our work is motivated by the emergence of AR and ``open'' data. In AR applications, a user can interact with what he/she sees on a screen, which enables manual annotation of street-views. On the other hand, several national and international open data initiatives, dedicated to the description of  territories, exist, leading to databases of semantic information at different scales (city-scale up to world-scale), for example: OpenStreetMap\footnote{OpenStreetMap: https://www.openstreetmap.org/} and Mapillary\footnote{Mapillary: https://www.mapillary.com}. Additionally, thanks to the evolution of territorial policies, more and more national mapping agencies also make available different thematic layers of their maps, which contain abundant semantic information. All these databases are regularly updated and represent a rich information source that can be linked to multimedia data, but they are currently under-exploited for visual localization.

The proposed semantic signature can be used in two ways: 1) as an individual localization method, it achieves coarse localization; 2) as a complement to other localization methods, it can effectively reduce the search scope by filtering out irrelevant regions. In large-scale applications, coarse localization can be used as a preceding step before pose estimation. Given a query image, a sophisticated localization work flow might consist of the following steps (see Fig.~\ref{fig_application_scenario}):
\begin{enumerate}
\item Perform feature detection;
\item Narrow down the search scope using semantic features;
\item Retrieve relevant images or low-level visual features;
\item Perform place recognition or pose estimation.
\end{enumerate}
This paper only covers the second step in the above pipeline, which focuses on the representation, indexing, and matching of semantic information. We foresee that real applications of urban localization using street-view images captured by mobile devices will emerge in the near future.

\section{Related work}
\label{sect_related_work}

Existing work on visual localization can be mainly divided into two categories: feature point retrieval and image retrieval. In the former approach, place recognition and camera pose estimation are solved by point-based voting and matching. For example, Schindler et al. propose a city-scale place recognition scheme~\cite{Schindler2007}. They use a vocabulary tree~\cite{Nister2006} to index SIFT features with improved strategies for tree construction and traversal. 
Irschara et al.~\cite{Irschara2009} also use a vocabulary tree for sparse place recognition using 3D-point clouds. They not only use real views, but also generate synthetic views to extend localization capability.
Li et al.~\cite{Li2010} address city-scale place recognition and focus on query efficiency. They prioritize certain database features according to a set covering criterion, and use a randomized neighborhood graph for data indexing and approximate nearest neighbor search~\cite{Arya1993}.
Zamir and Shah~\cite{Zamir2010} use Google street-view images for place recognition. They distinguish single image localization and image group localization, and derive corresponding voting and post-processing schemes for refined matching. 
Chen et al.~\cite{Chen2011} study the localization of mobile phone images using street-view databases. They propose to enhance the matching by aggregating the query results from two datasets with different viewing angles. 
Sattler et al.~\cite{Sattler2011} propose to accelerate 2D-to-3D matching by associating 3D-points with visual words and prioritizing certain words. 
Zhang et al.~\cite{Zhang2011} address performance degradation in large urban environments by dividing the search area into multiple overlapping cells. Relevant cells are identified according to coarse position estimates by e.g. A-GPS and query results are merged.
Li et al.~\cite{Li2012} consider worldwide image pose estimation. They propose a co-occurrence prior based RANSAC~\cite{Fischler1981} and bidirectional matching to maintain efficiency and accuracy.
Lim~\cite{Lim2012} et al. address real-time 6-DOF estimation in large scenes for auto-navigation. They use a dense local descriptor DAISY~\cite{Tola2008} instead of SIFT for fast key-point extraction, and a binary descriptor BRIEF~\cite{Calonder2010} for key-point tracking. 

The other category of localization techniques is based on image retrieval. Conventionally, this is only used for place recognition~\cite{Crandall2009,Shrivastava2011}. For example, Zamir and Shah~\cite{Zamir2014} propose multiple nearest neighbor feature matching with generalized graphs. Arandjelovic and Zisserman~\cite{Arandjelovic2014} propose an improved bag-of-features model. Torii et al.~\cite{Torii2015} apply the VLAD descriptor~\cite{Jegou2010} to synthesis views. 
Arandjelovic et al.~\cite{Arandjelovic2016} extend VLAD with a deep neural network architecture to address  scene appearance changes due to long-term acquisitions, day/night or seasonal changes. Iscan et al.~\cite{Iscen2017} propose to aggregate descriptors from panoramic views.
Since 3D-point datasets can be built from 2D images with structure-from-motion techniques~\cite{Agarwal2009}, it is possible to directly estimate 6-DOF with an image database. Recently, Song et al.~\cite{Song2016} propose to estimate 6-DOF after image retrieval. Sattler et al.~\cite{Sattler2017} show experimentally that image retrieval approaches are perhaps more suitable for large-scale applications.
With the success of deep learning, more approaches based on learned features are proposed to highlight effective visual features and exploit multiple modalities. A recent survey can be found in \cite{Piasco2018}.


Our approach is different from existing work, because we use semantic information instead of visual information. A relevant idea can be found in~\cite{Ardeshir2014}, where Ardeshir et al. use existing knowledge of objects to assist object detection. While they show the potential of semantic objects in localization, we perform more extensive study in this paper. We also find that edit distance works better than their histogram based metric.
Another related scheme is~\cite{Arth2015}, where Arth et al. use a different kind of semantic information. They perform re-localization by extracting straight line segments of buildings from a query image and comparing with a database. While our work focuses on objects, it can also be extended to include other semantic features, such as building corners~\cite{Arth2015}.
On the other hand, our approach can also be used as an initial step to narrow down the search scope for some existing work, such as~\cite{Song2016,Bhowmik2017,Sattler2017}.

\section{The proposed scheme}
\label{sect_proposed}

\begin{figure}
\centering
\includegraphics[scale=0.4]{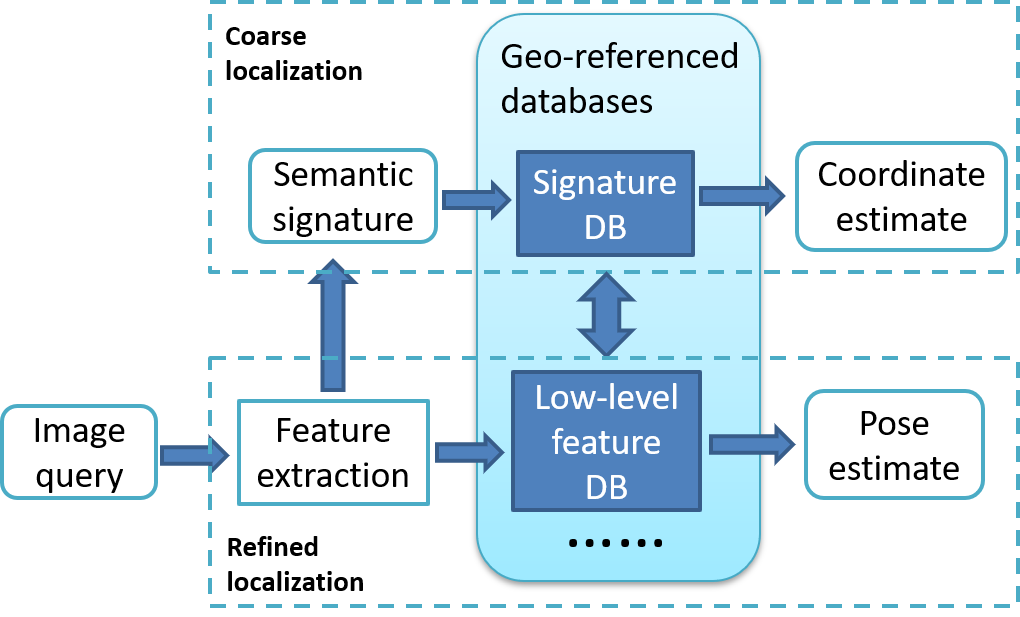}
\caption{A complete application scenario of visual localization. This paper only focuses on the upper path.}
\label{fig_application_scenario}
\end{figure}

The target application is localization in urban environments.
In a typical scenario, a user has a mobile device that captures images of the surrounding area. The goal is to tell the user's location according to these images. In a retrieval-based approach, it is tackled by extracting information from the images and comparing with a geo-referenced database. 
Figure~\ref{fig_application_scenario} illustrates a complete application scenario, which is divided into coarse localization and refined localization.
A critical question there is what kind of information to extract from images. In this work, the focus is on semantic information, which corresponds to the upper path in Fig.~\ref{fig_application_scenario}. Semantic information is high-level information based on human perception. In our context, it is about what people see from images. For example, people can tell their location by describing their surroundings. The same principle can be applied to localization. Since the images taken by the mobile device are typically street views, the semantic information contains objects such as buildings, streets, the sky, the ground, cars, humans, etc. It is found that some of these objects are useful for localization. In general, \textit{semantic objects} with the following properties are of particular interest:
\begin{itemize}
\item Permanent -- the object does not move;
\item Informative -- the object is distinguishable from others;
\item Widely available -- the object is distributed in the scene.
\end{itemize}
Additionally, the objects should have unambiguous locations and be suitable for object detection algorithms. 
In this paper, we assume that detecting such objects is feasible and focus on retrieval aspects.

\subsection{Semantic signatures}
\begin{figure}
\centering
\includegraphics[scale=0.4]{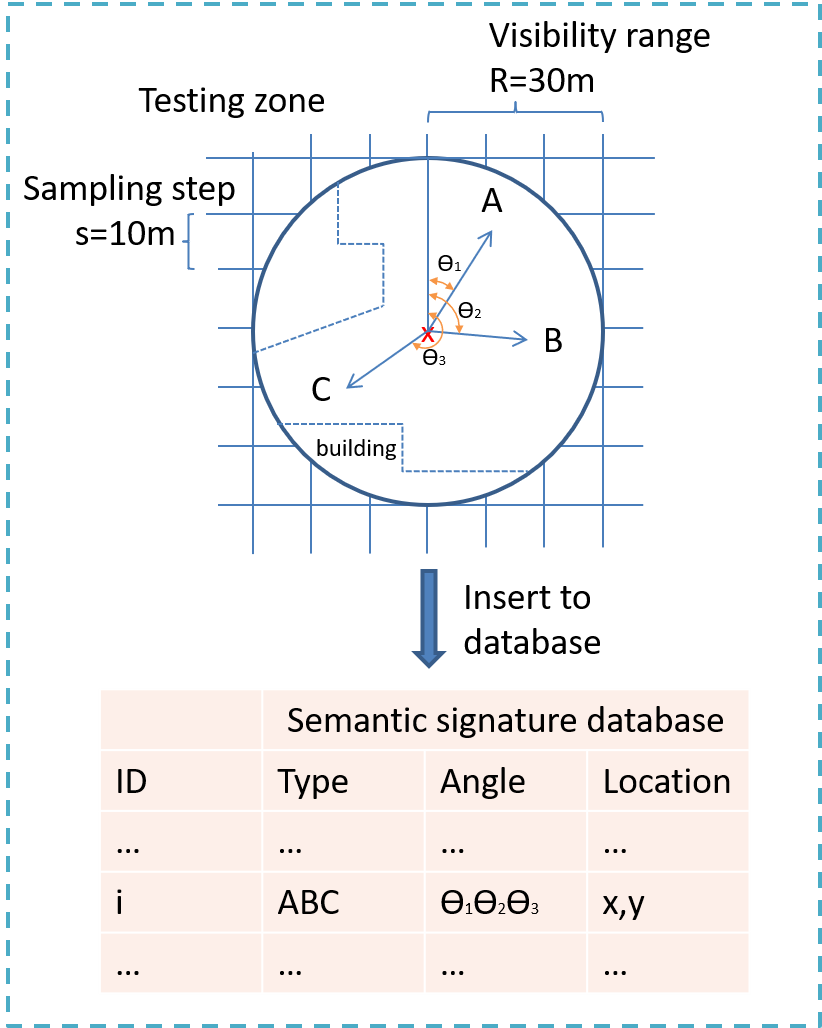}
\caption{The generation of semantic signatures.}
\label{fig_semantic_signature}
\end{figure}

\begin{figure}
\centering
\begin{tabular}{ccc}
\includegraphics[scale=0.02]{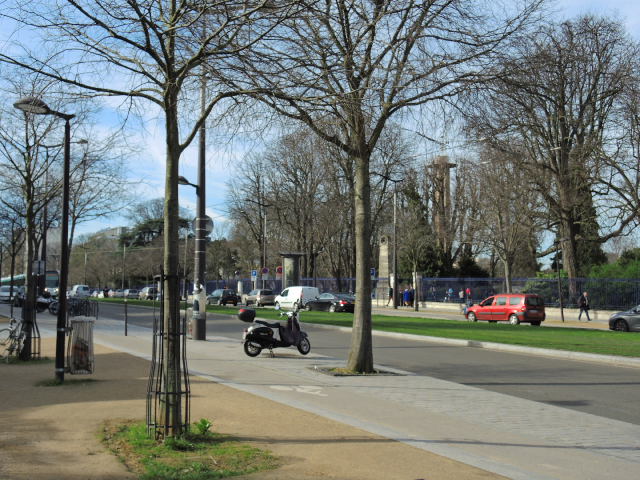} &
\includegraphics[scale=0.02]{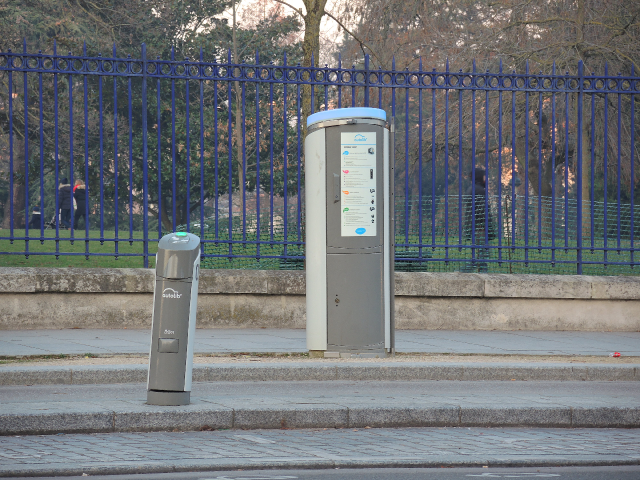} &
\includegraphics[scale=0.02]{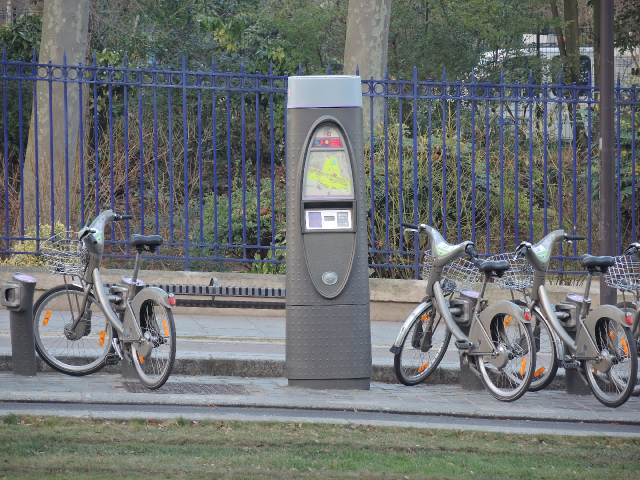} \\
alignment tree (B) & autolib station (J) & bike station (H)\\
\includegraphics[scale=0.02]{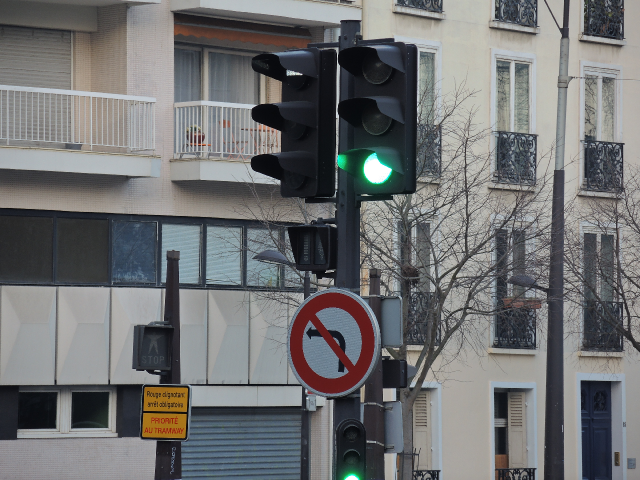} &
\includegraphics[scale=0.02]{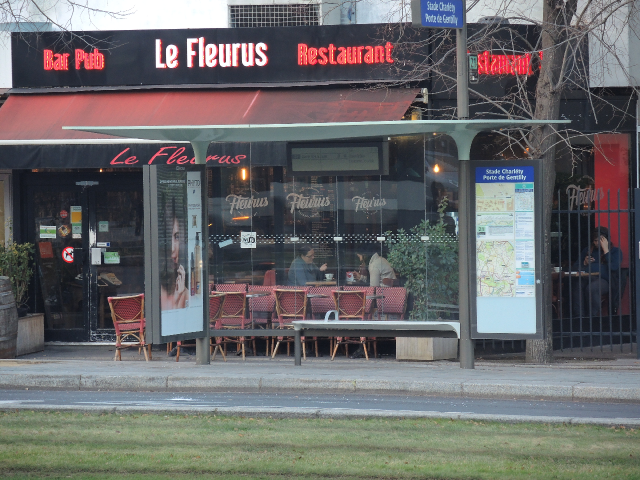} &
\includegraphics[scale=0.02]{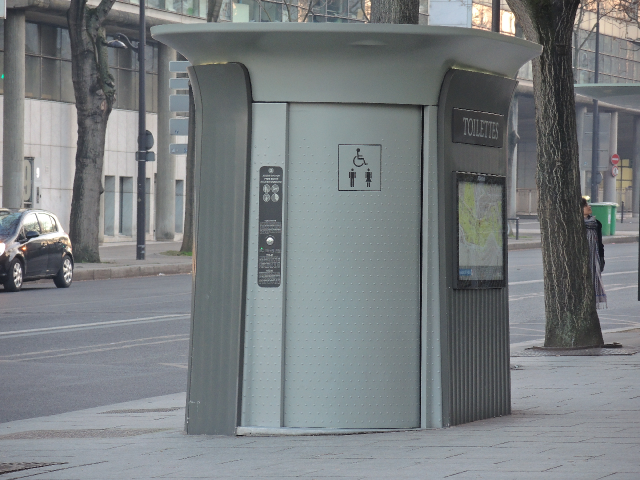} \\
traffic light (G) & bus stop (M) & automatic WC (I)
\end{tabular}
\caption{Some examples of semantic objects (see Table~\ref{tab_semantic_objects}).}
\label{fig_semantic_object}
\end{figure}

\begin{figure}
\centering
\includegraphics[scale=0.18]{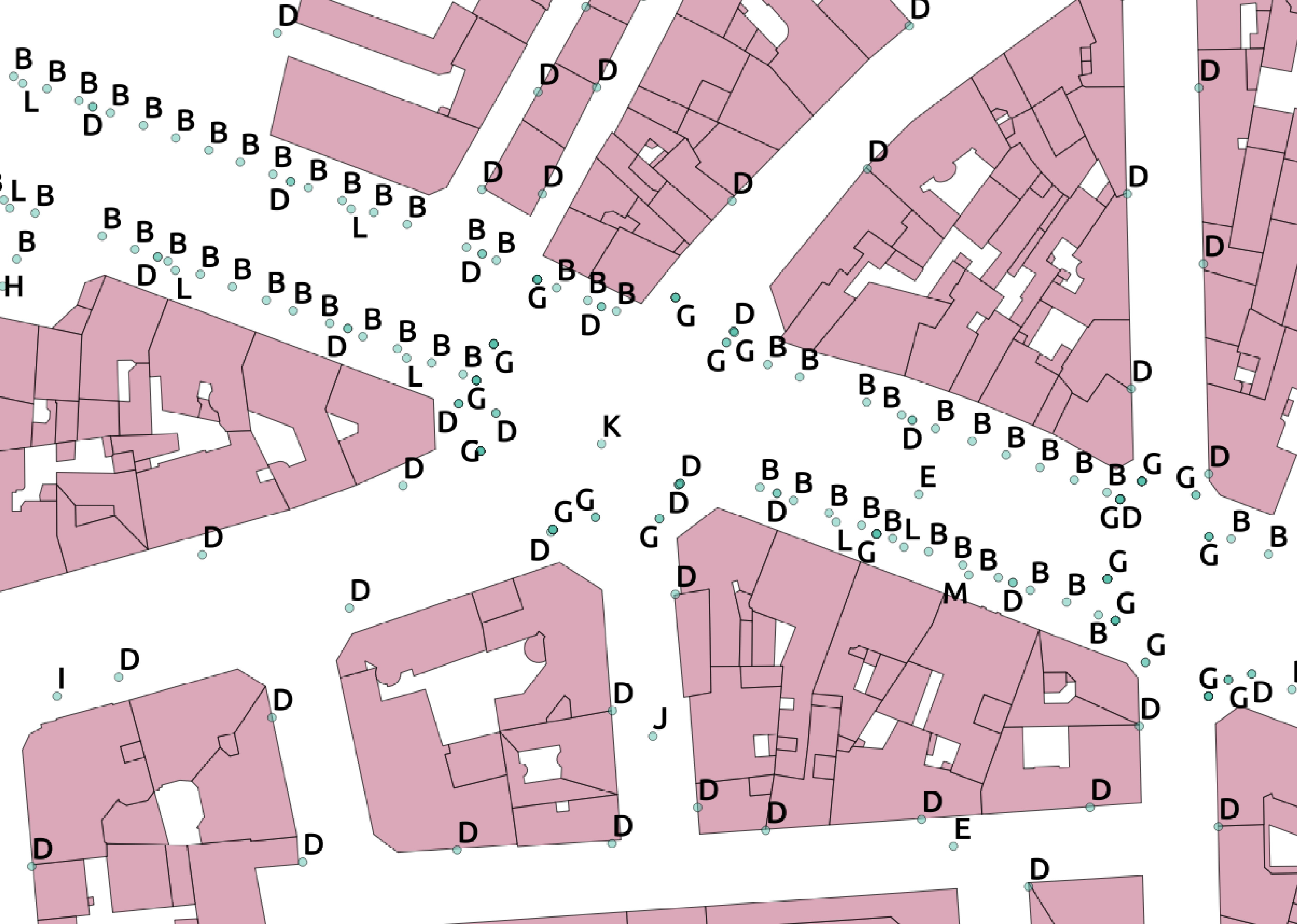}
\caption{Distribution of semantic objects on Paris streets.}
\label{fig_semantic_object_distribution}
\end{figure}

Once semantic objects have been detected, they are encoded into a compact representation, which we call \textit{semantic signature}. A semantic signature describes some properties of the corresponding objects. It is required to be compact and easily indexable. In this work, we propose to compose a semantic signature by:
\begin{itemize}
\item Object type -- the category (class) of an object;
\item Object angle -- the relative angle of an object.
\end{itemize}
Specifically, the object type is a label, denoted by $t$; the object angle is measured according to the north and a view point, denoted by $a$. Given a view point coordinate $(x,y)$ and a visibility range $R$, each location can be associated with a semantic signature, which is related to the semantic objects that can be seen from that location. In our implementation, semantic objects are identified by a panoramic sweep in a clockwise order starting from the north. A semantic signature is the concatenation of two parts: $s=\{s^{(1)}|s^{(2)}\}$, where $s^{(1)}=t_1|\cdots|t_n$ represents the type sequence of the corresponding objects, $s^{(2)}=a_1|\cdots|a_n$ represents the corresponding angle sequence, and $n$ is the number of visible semantic objects within $R$. Figure~\ref{fig_semantic_signature} illustrates the generation of semantic signatures. Some examples of semantic objects and their distribution are shown in Fig.~\ref{fig_semantic_object} and Fig.~\ref{fig_semantic_object_distribution} (see Table~\ref{tab_semantic_objects} for a complete list).
Ideally, each signature is unique, so that localization can be achieved by matching a query signature with a signature database. A database of semantic signatures can be built from existing data sources, such as geographic information systems.

In addition, it is required by one of our signature comparison metrics that the north is known when generating a signature. This is not unrealistic, because nowadays mobile devices are typically equipped with a compass.
The centroid of an object is used for representing its location. In order to have a stable angle sequence, it is necessary to quantize angle values. We use 4-bit quantization, i.e., each angle value is quantized by 16 levels ($22.5^\circ$ per step).

\subsection{Signature comparison}
Given two semantic signatures, an important question is how to compare them. Since localization is achieved by signature search and retrieval, a similarity metric is needed. Since a signature has two parts, for simplicity it is preferable to use a metric that is compatible with both parts. This is possible if the two parts are considered as two general sequences. In this work, we use the following metrics:
\begin{itemize}
\item Jaccard distance;
\item Histogram distance;
\item Edit distance.
\end{itemize}
Denote two ordered sequences as $\mathbf{x}=x_1\cdots x_n$, $\mathbf{y}=y_1 \cdots y_m$. The Jaccard distance~\cite{Jaccard1912} is defined as
\begin{equation}
\frac{|\mathbf{x} \cap \mathbf{y}|}{|\mathbf{x} \cup \mathbf{y}|} \ .
\end{equation} 
The histogram distance is defined as
\begin{equation}
\sum_c{\frac{\mbox{min}\{|\mathbf{x}_c|,|\mathbf{y}_c|\}}{\mbox{max}\{|\mathbf{x}_c|,|\mathbf{y}_c|\}}} \ .
\end{equation}
where $c$ represents an object class. This metric was used in~\cite{Ardeshir2014}, so it is a good candidate for performance comparison.
The edit distance (a.k.a. the Levenshtein distance)~\cite{Navarro2001} is defined by the recurrence

\begin{align}
d_{i0} & =\sum_{k=1}^i{w_{del}(y_k)} \ , \   d_{0j}=\sum_{k=1}^j{w_{ins}(x_k)}   \nonumber \\
d_{ij} & = 
\begin{cases}
d_{i-1,j-1} & \text{if } x_j=y_i \\
\text{min} 
\begin{cases}
d_{i-1,j}+w_{del}(y_i)\\
d_{i,j-1}+w_{ins}(x_j)\\
d_{i-1,j-1}+w_{sub}(x_j,y_i)
\end{cases}
& \text{if } x_j \neq y_i
\end{cases} \nonumber
\end{align}
where $1\leq i \leq m$, $1\leq j \leq n$, and $w_{del}, w_{ins}, w_{sub}$ (set to $1$ by default) are the weight factors for deletion, insertion and substitution respectively. This metric requires that the north is known when generating signatures.

Given two sequences of symbols, these metrics compare the value or the order of the symbols, but they exhibit different levels of ``strictness''. The Jaccard distance only considers the occurrence and completely ignores the order and the frequency; the histogram distance also ignores the order but counts the frequency of symbols; the edit distance takes into account both the order and the frequency. By selecting different metrics, different trade-offs between robustness and discrimination power can be achieved. A coarse metric is useful for rough and quick matching, while a fine-grained metric is useful for refined matching. On the other hand, the computation cost is also different. The more complex the metric, the more computation.

\subsection{Retrieval schemes}
The localization problem is solved by a retrieval-based framework. The procedure starts with a panoramic query image captured by a mobile device. Then the following basic steps apply:
\begin{enumerate}
\item A query signature is computed from the query image;
\item Similar signatures are retrieved from a database according to the query signature;
\item The best $t$ matches are returned. 
\end{enumerate}
After the best matches are identified, post-processing schemes may follow depending on the specific application. 
In this paper, the focus is to find the best matches in an accurate and efficient way. Since a semantic signature has two parts -- type and angle, an essential question is how each part contributes to the similarity between two signatures. In order to facilitate different occasions where one may choose to favour accuracy or efficiency, we propose two retrieval methods: ``metric fusion'' and ``two-stage metric fusion''. They are specially designed for our scenario, but have some resembelance to the concepts of early fusion and late fusion in content classification~\cite{Snoek2005} and retrieval~\cite{Vrochidis2019}.

\subsubsection{Metric fusion}
\begin{figure}
\centering
\begin{tabular}{cc}
\includegraphics[scale=0.35]{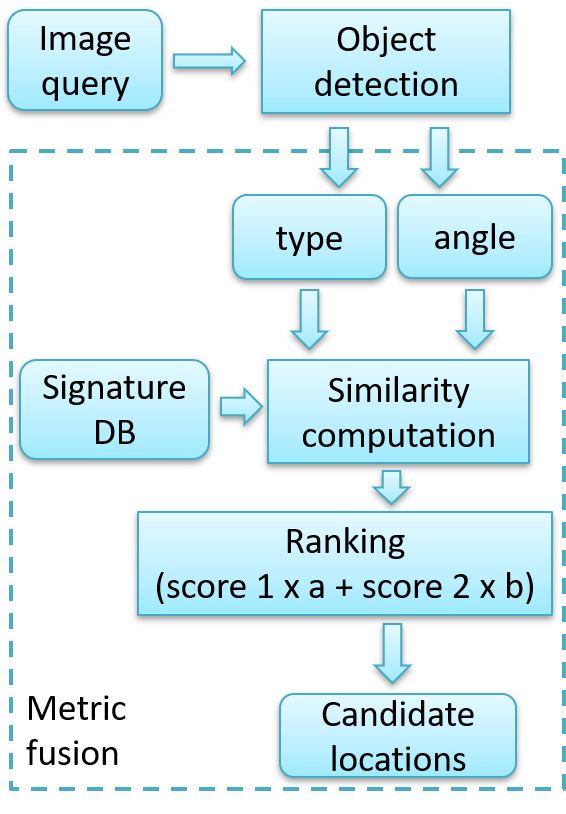} &
\includegraphics[scale=0.35]{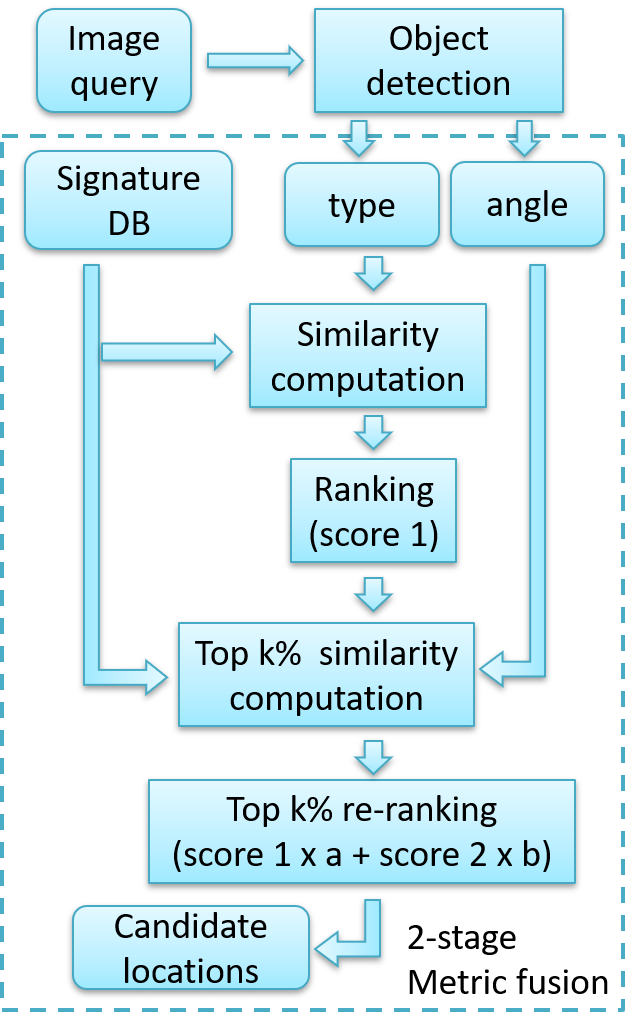} \\
(a) & (b)
\end{tabular}
\caption{Metric fusion (a) and two-stage metric fusion (b).}
\label{fig_metric_fusion}
\end{figure}

In this method (see Fig.~\ref{fig_metric_fusion}a), a similarity score is first computed from each part of the signature. Then a weighted sum of the two scores is computed. Signatures are ranked according to the total score. Denote two signatures as $s_a=\{s_a^{(1)}|s_a^{(2)}\}$ and $s_b=\{s_b^{(1)}|s_b^{(2)}\}$. The distance $d$ is defined as
\begin{eqnarray}
d & = & \alpha \cdot d_1+ \beta \cdot d_2 \\ \label{eqn_metric_fusion}
  & = & \alpha \cdot D_1(s_a^{(1)},s_b^{(1)}) + \beta \cdot D_2(s_a^{(2)},s_b^{(2)})
\end{eqnarray}
where $\alpha$ and $\beta=1-\alpha$ are weight factors, $D_1$ and $D_2$ are the chosen similarity metrics.
By maintaining sufficiently large weight factors, both type and angle information is aggregated. When $\alpha$ or $\beta$ is zero, the scheme reduces to single metric based ranking.

\subsubsection{Two-stage metric fusion}
A drawback of the previous scheme is heavy computation: two metrics are computed for each pair of signatures. Although this approach might give the most accurate ranking, in practice only the top ranks are useful. Therefore, it is possible to improve the speed. Intuitively, if two signatures do not match, then any metric is likely to result in a low rank.
An improved scheme works as follows (see Fig.~\ref{fig_metric_fusion}b):
\begin{enumerate}
\item A similarity score is first computed from one part of the signature;
\item Top $k\%$ candidate locations are retrieved;
\item For the retrieved candidates, another score is computed from the other part of the signature, and added to the previous score as in Eqn.~\eqref{eqn_metric_fusion};
\item The top $k\%$ candidates are re-ranked according to the new score.
\end{enumerate}
By adjusting $k$, different trade-offs between accuracy and efficiency can be achieved. When $k=100$, the scheme becomes the standard metric fusion.

\subsection{Prerequisite and post-processing}
The proposed scheme utilizes two properties of semantic objects -- type and angle. In general, an object recognition algorithm is needed to provide such information. State-of-the-art candidate algorithms are typically based on region proposals and convolutional neural networks, such as~\cite{Girshick2015,He2015,Ren2015}. 
In case an object recognition algorithm is not available, the type information can also be provided by a human user (because the semantic objects are easy to recognize) as a query, which is an alternative way to use the proposed scheme. Experiment results later show that even if angle information is missing, type information can individually facilitate localization, and vice versa.

The general goal of the proposed scheme is to provide a list of potential locations according to a query signature. How to derive a final answer from the candidate locations is the task of post-processing. This is not the focus of the paper, but we briefly discuss some particular procedures in the following.

If no further processing is desired, a most straightforward way is to take the best match as the answer, i.e., $t=1$. When $t>1$, some analysis can be performed with the candidate locations. For example, it is possible to narrow down the search range by obtaining some prior information about the ``popularity'' of locations -- some locations are more likely to be visited than others. If extra information is available, such as street-view images or 3D models at the candidate locations, then one may perform 2D-to-2D~\cite{Song2016,Bhowmik2017} or 2D-to-3D~\cite{Sattler2011} matching using the query image. However, since these operations are expensive in computation and data storage, it is desirable to restrain them in a small scale.
Therefore, it is important that the proposed scheme returns ``good'' candidates in a short list. This is confirmed by the experiment results.

\section{Experiments}
\label{sec:experiments}

The proposed scheme has been extensively evaluated with a city-scale dataset. The dataset, the evaluation framework, and the results are presented in this section.

\subsection{The dataset}
\begin{figure}
\centering
\begin{tabular}{cc}
\includegraphics[scale=0.25]{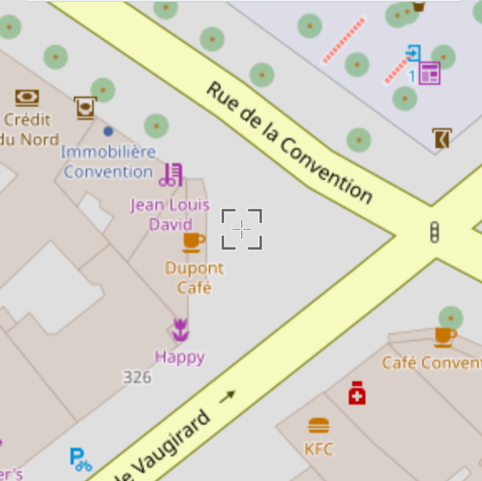} &
\includegraphics[scale=0.25]{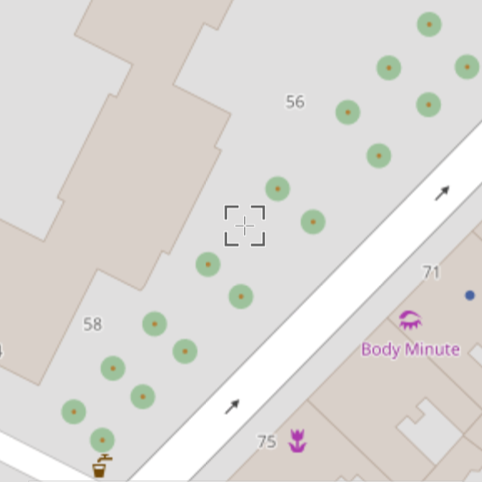} \\
(a) 2.2961483$^\circ$ 48.8372312$^\circ$  & (b) 2.3181783$^\circ$ 48.8299839$^\circ$\\
\includegraphics[scale=0.25]{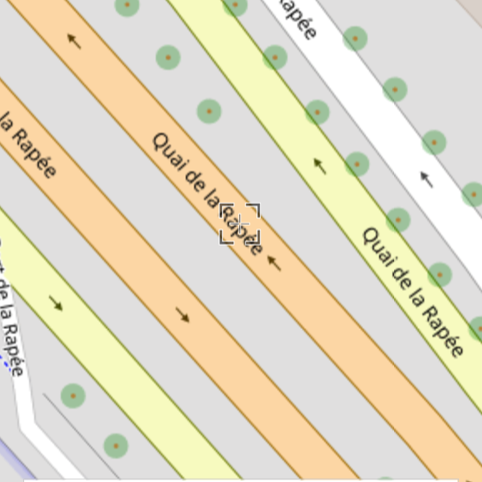} &
\includegraphics[scale=0.25]{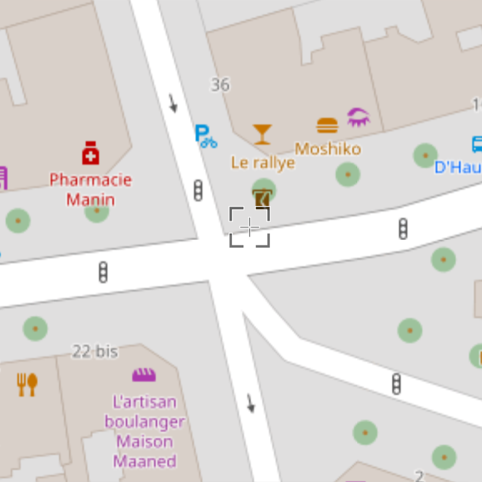}  \\
(c) 2.3714248$^\circ$ 48.8426869$^\circ$ & (d) 2.3879806$^\circ$ 48.8832431$^\circ$
\end{tabular}
\caption{Some examples of successful query locations (longitude/latitude). Their signatures contain mostly alignment trees and street lights.}
\label{fig_query_examples_good}
\end{figure}

\begin{figure}
\centering
\begin{tabular}{cc}
\includegraphics[scale=0.25]{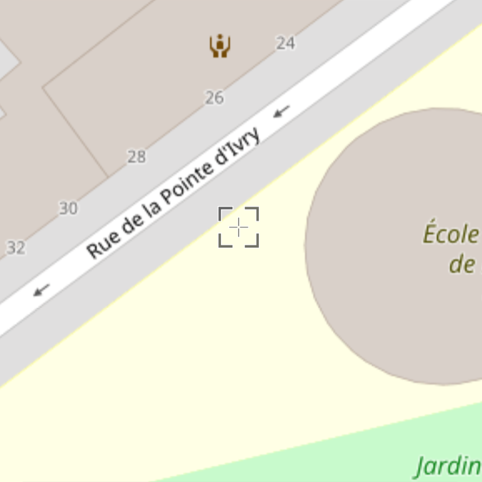} &
\includegraphics[scale=0.25]{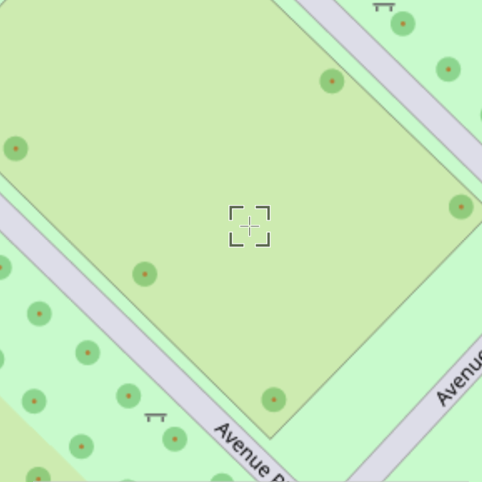} \\
(a) 2.3639067$^\circ$ 48.8223214$^\circ$  & (b) 2.2998637$^\circ$ 48.8547905$^\circ$\\
\includegraphics[scale=0.25]{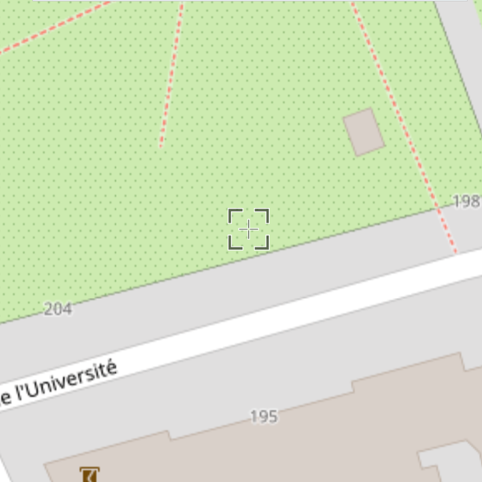} &
\includegraphics[scale=0.25]{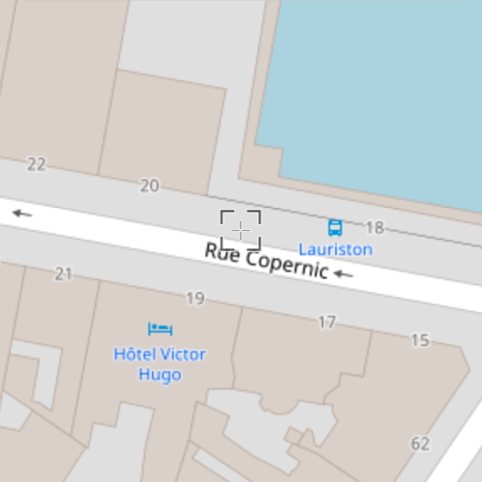}  \\
(c) 2.2991023$^\circ$ 48.8607219$^\circ$ & (d) 2.2893088$^\circ$ 48.869208$^\circ$
\end{tabular}
\caption{Some examples of unsuccessful query locations (longitude/latitude). Their signatures only contain a few street lights.}
\label{fig_query_examples_bad}
\end{figure}

Our dataset is about Paris. It consists of approximately $300,000$ semantic signatures that cover most of the city. These signatures are built from $11$ categories of objects, as listed in Table~\ref{tab_semantic_objects}.
\begin{table}
\centering
\caption{Semantic objects.}
\begin{tabular}{llll}
\hline
ID & Name & Number & Symbol \\
\hline
1 & Alignment tree & 1752696 & B \\ 
2 & Water fountain & 6713 &  C   \\
3 & Street light   & 2299639 & D \\ 
4 & Indicator      & 36333   & E \\
5 & Traffic light  & 102240  & G \\
6 & Bike station   & 14397   & H \\
7 & Automatic WC   & 8006    & I \\
8 & Autolib (car) station & 4421   & J \\
9 & Taxi station   & 2537    & K \\
10 & Public chair  & 135748  & L \\
11 & Bus stop      & 32320   & M \\
\hline
\end{tabular}
\label{tab_semantic_objects}
\end{table}
These objects are found from Open Data Paris\footnote{Open Data Paris (\url{https://opendata.paris.fr}) hosts a collection of more than $200$ public datasets provided by the city of Paris and its partners.} with known coordinates. 
The signature database is constructed by sampling the Paris region with a step of $s=10$ meters. At each sampling point (cell), a semantic signature is created to summarize objects within 30 meters, i.e., the visibility range $R$ is set to 30.
Some basic properties of the database are listed in Table~\ref{tab_signature_database}a. Each database record contains a location (represented by a $s^2=100\text{m}^2$ cell) and its signature. If database records are grouped by the signature using only type information, then the number of groups is approximately $45\%$ of the number of signatures (see Table~\ref{tab_signature_database}b), i.e., on average less than three cells have the same signature. It is expected that each signature group contains only one cell. Some more statistics about the signature groups are listed in Table~\ref{tab_signature_database}b. It is true that most signature groups ($\geq 75\%$) have only one cell. This is crucial to effective localization. Note that there are also rare cases when it is almost impossible to find the correct location. For example, there are $29958$ cells with the same signature type ``DDD'', which means three street lights.
This implies that our proposed solution works in a probabilistic sense.
In general, whether a location query will be successful depends on the entropy of its signature. Table~\ref{tab_signature_database}a gives the average length of a signature. If a signature is longer than the average and contains multiple object types, it is likely to be effective, and vice versa. 
Some examples of successful and unsuccessful query locations are shown in Fig.~\ref{fig_query_examples_good}--\ref{fig_query_examples_bad}.
Nevertheless, the localization power can be improved when type and angle information is combined. The last column of Table~\ref{tab_signature_database}b shows that angle information is even more discriminative than type information.
It is also worth noting that the overall file storage only takes 38.7 MB (without optimization) to cover a large area. This is an extremely small cost for city-scale localization. Conventional low-level feature based approaches, e.g.~\cite{Li2012,Bhowmik2017}, at least require several GBs even for a small scene.

\begin{table}
\centering
\caption{Database properties.}

\begin{tabular}{cc}

(a) basic properties & (b) signature group size \\

\begin{minipage}{0.5\linewidth}
\centering

\begin{tabular}{p{2.6cm}l}
\hline
Visibility range & 30 meters\\
No. of signatures & 312134\\
Mean signature length & 14 objects\\
Covered area & 79 $\text{km}^2$ \\
Data storage & 38.7 MB \\
\hline
\end{tabular}

\end{minipage} 

&

\begin{minipage}{0.5\linewidth}
\centering
\begin{tabular}{lll}
\hline
         & by type &  by angle \\
\hline
count    & 140296 & 204891 \\
mean     &     2.2 & 1.5\\
std      &   121.5 & 11.1\\
min      &     1  & 1\\
25\%      &    1  & 1\\
50\%      &     1 & 1\\
75\%      &     1 & 1\\
max      & 29958 & 1240\\
\hline
\end{tabular}

\end{minipage}

\end{tabular}
\label{tab_signature_database}
\end{table}

For large-scale retrieval applications, it is necessary to consider database indexing schemes for efficiency. Nevertheless, for the demonstration in this work, it suffices to use a linear scan scheme for signature retrieval, thanks to the compactness of signatures. Since a signature is encoded by symbols of small alphabets, more efficient indexing is possible if necessary. For example, signatures can be clustered and indexed according to certain patterns. A natural way to generate a pattern is to gather distinct symbols in a signature and sort them, which is actually the representation used by Jaccard distance.

\subsection{The evaluation framework}
\label{sect_evaluation_framework}
The signature database can be matched with other signatures obtained by various means. In order to evaluate the retrieval aspects of the proposed scheme, we skip object detection. A query set is formed by randomly selecting ten thousand locations and the associated signatures from the database. Each signature in the query set is used for querying the database. The average performance for all queries is noted. We mainly consider two benchmarks:
\begin{itemize}
\item Cumulative distribution of distance errors;
\item Recall rate of correct locations.
\end{itemize}
The first benchmark measures the average distance from the ground truth location to a candidate location. The best results among top $t$ candidates is noted. In our experiments, we set $t=100$. The second benchmark examines the rank of the ground truth location among all candidates, emphasizing the capability as a filtering tool. It can be considered as a special retrieval scenario with only one relevant answer per query. They will be explained with more details later.

\subsubsection{Distortion simulation}
In practice, object detection is not perfect. Using the query set directly does not reveal the performance in reality. Therefore, we propose to simulate errors in object detection. The simulated  operations are listed in Table~\ref{tab_distortion_simulation}.
\begin{table}
\caption{Simulated signature distortion.}
\centering
\begin{tabular}{lll}
\hline
ID & Distortion type  & Comment\\
\hline
1 & Miss detection & Remove objects \\
2 & False detection & Introduce new objects\\
3 & False classification & Change object type\\
4 & Angle noise & Add noise to each angle\\
\hline
\end{tabular}
\label{tab_distortion_simulation}
\end{table}
In a more complete setting, each query item is first randomly distorted before matching with the database. We consider three levels of distortion -- light, medium, and strong, corresponding to 1, 7, or 13 occurrences of random distortion, including miss detection, false detection, and false classification. Each time up to more than $50\%$ objects in a signature are distorted. In addition to type distortion, angle noise is always applied following a normal distribution with the standard deviation equal to $5$ and the maximum value clipped to $30$. The distortion parameters are set empirically. They serve as guidelines if upstream visual processing tools are to be designed for our application.
 
In reality, the queries might be chosen off-grid, which corresponds to distance and angle changes from the nearest sampling points. These effects are also simulated by the distortions to some extent.

\subsection{Experiment results}
In this section, we evaluate the proposed scheme in terms of the two benchmarks defined in Sect.~\ref{sect_evaluation_framework}. According to the average signature length (14 objects), we mainly consider performance under medium level distortion. The two signature parts, type and angle, are separately tested first, followed by metric fusion and two-stage metric fusion schemes. In addition, the computation complexity is also measured. After extensive tests, practical configurations are identified and further examined with various system parameters.

\subsubsection{Localization performance}
\begin{figure}
\centering
\includegraphics[scale=0.4]{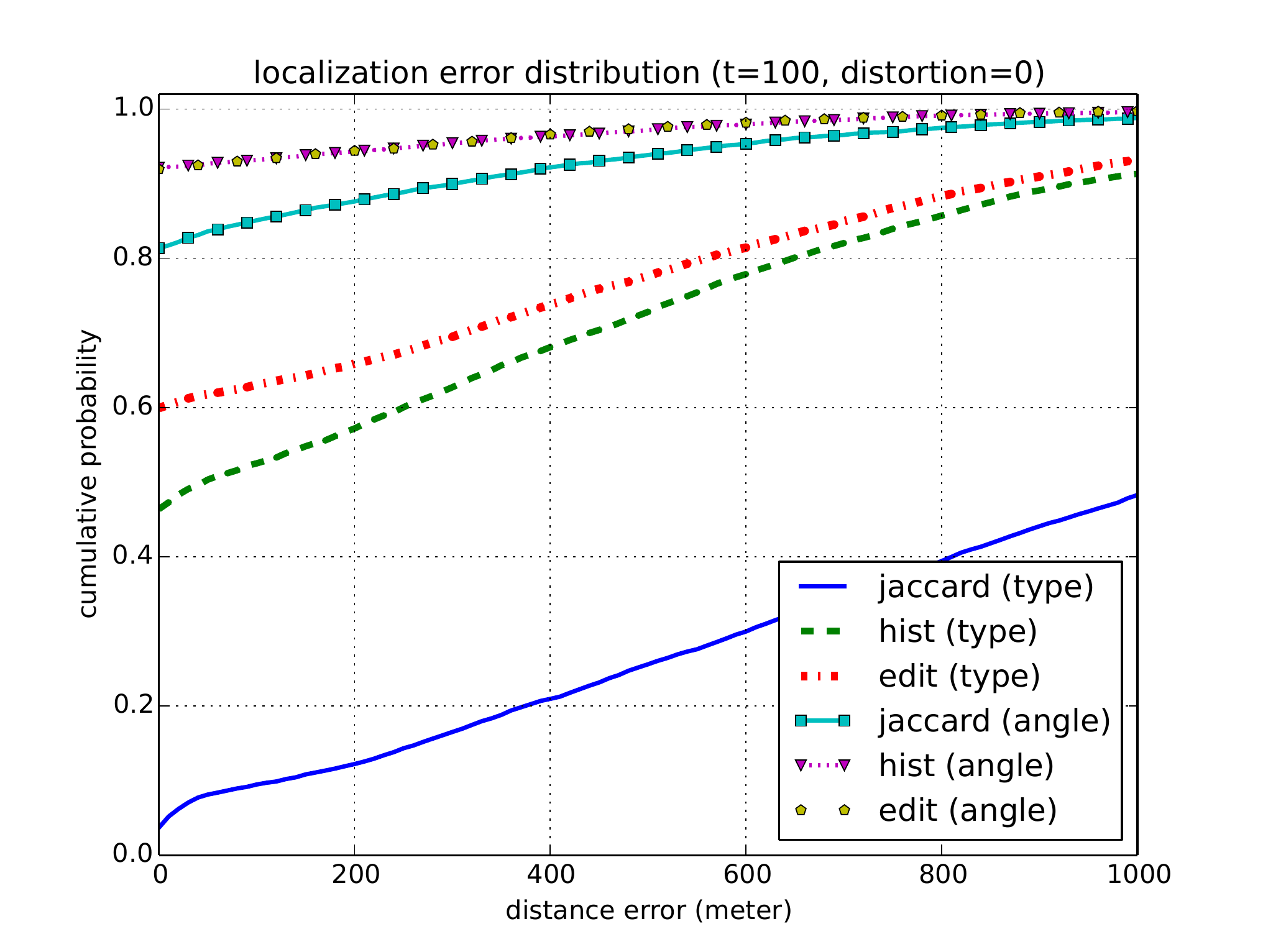}
\caption{Localization error (single metric, no distortion).}
\label{fig_localization_single_metric_t100_d0}
\end{figure}

\begin{figure}
\centering
\includegraphics[scale=0.4]{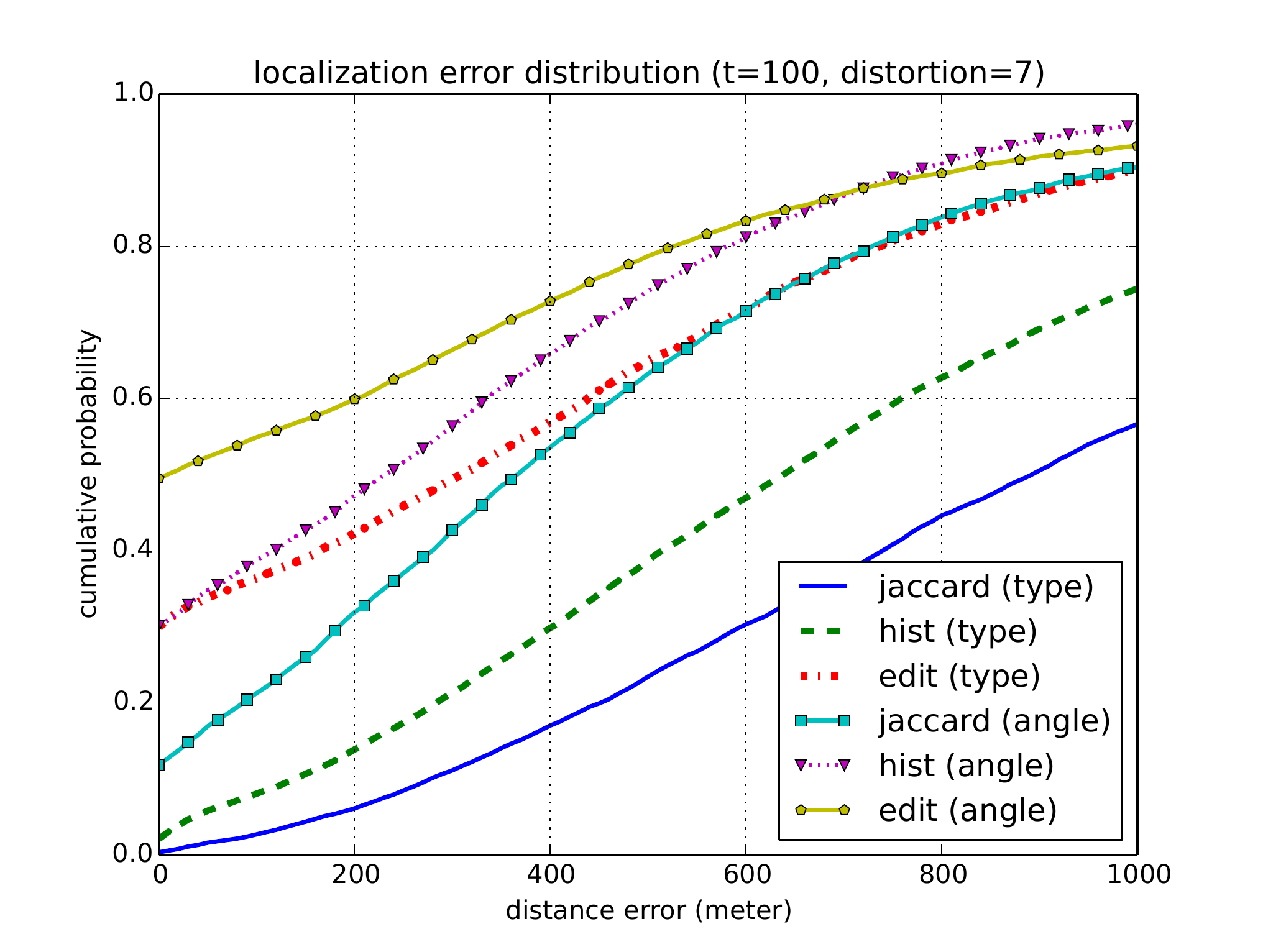}
\caption{Localization error (single metric, medium distortion).}
\label{fig_localization_single_metric_t100_d7}
\end{figure}

We first examine the effectiveness of the signature scheme. Figure~\ref{fig_localization_single_metric_t100_d0} shows the ideal localization error when only one part of a signature is used without distortion. There are six curves corresponding to the three metrics and the two signature components. For each point $(x,y)$ on a curve, it means for $y\times 100\%$ queries, the distance error is not larger than $x$. In general the cumulative probability increases with the localization error. A higher curve means better performance. We observe that both type and angle information can be used for localization, but the smaller error shows that angle generally works better. Among the three metrics, edit distance is the best, followed by histogram distance and Jaccard distance. In the best case, i.e. perfect object detection, more than $90\%$ of queries result in the correct location or have errors less than 10 meters, which is close to GPS accuracy.
When there is distortion, a similar trend can be observed in Fig.~\ref{fig_localization_single_metric_t100_d7}. The localization error increases with the distortion. The maximum query percentage for no error drops to $50\%$. But edit distance still performs the best.

\begin{figure}
\centering
\includegraphics[scale=0.4]{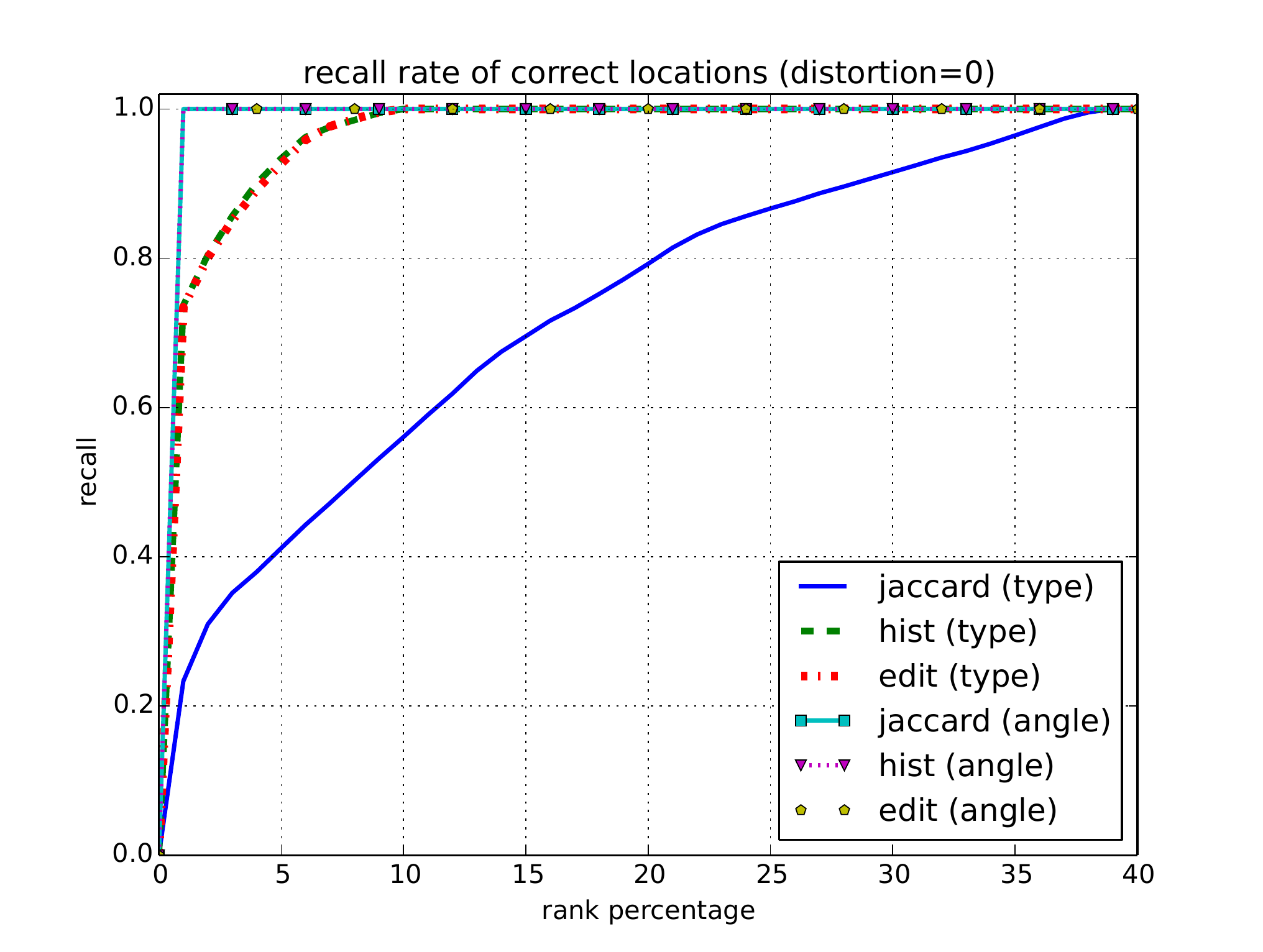}
\caption{Location recall (single metric, no distortion).}
\label{fig_groundtruth_single_metric_d0}
\end{figure}

\begin{figure}
\centering
\includegraphics[scale=0.4]{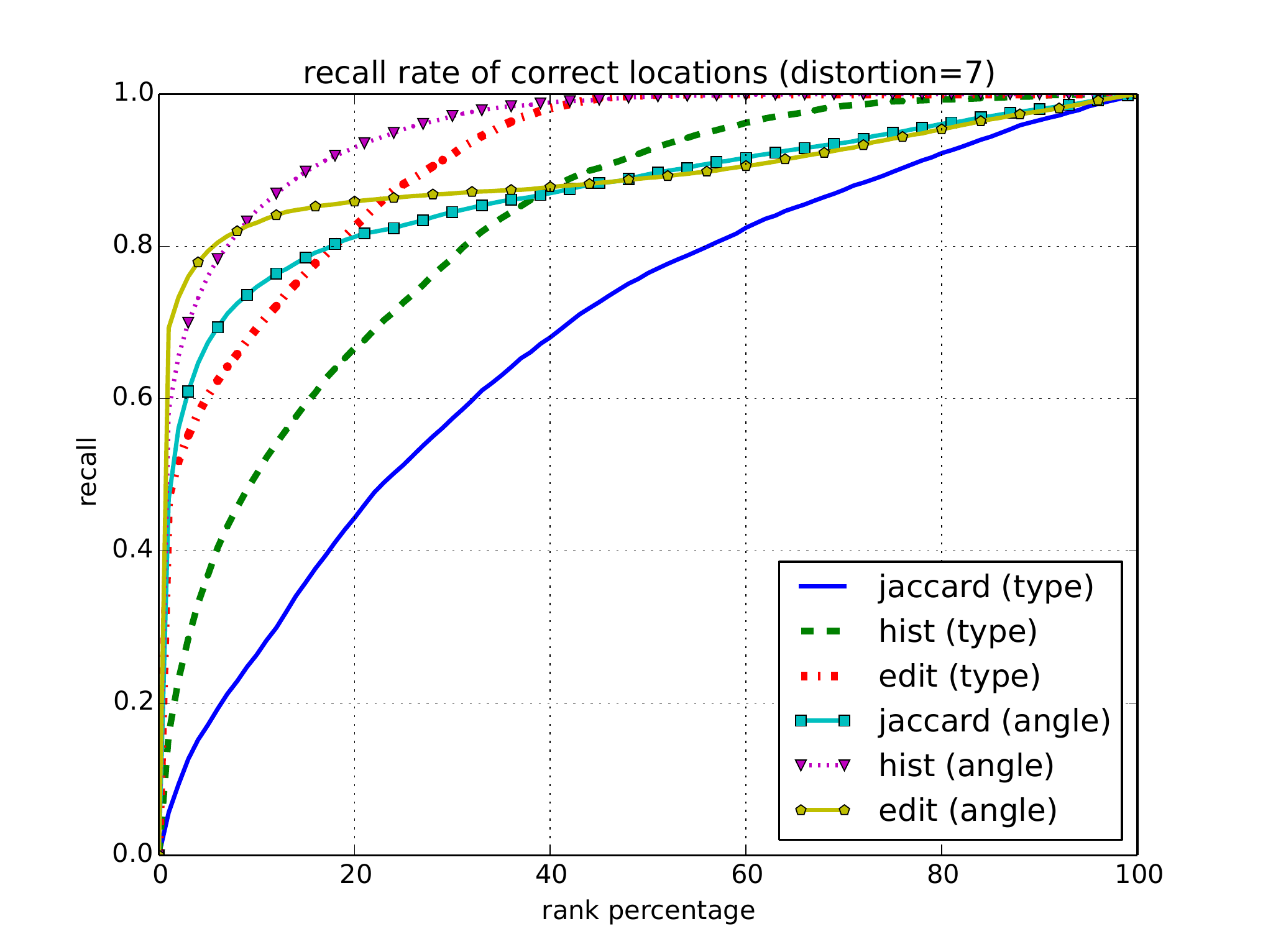}
\caption{Location recall (single metric, medium distortion).}
\label{fig_groundtruth_single_metric_d7}
\end{figure}

Next, we consider another benchmark -- the recall rate. In Fig.~\ref{fig_groundtruth_single_metric_d0}, a point $(x,y)$ means for $y\times 100\%$ queries, the corresponding ground truth rank is not lower than $x\%$. Ideally, we expect the recall to be as high as possible. The results show that given a query, our proposed method can effectively filter out irrelevant regions. For example, almost in all cases (except for Jaccard distance with type information), all queries have recall@10\%=$1$. That means only the top $10\%$ database candidates need to be considered. In Fig.~\ref{fig_groundtruth_single_metric_d7}, the recall is plotted for medium distortion. Although there is some performance degradation, the good settings can still keep the ground truth rank within top $20\%$ for $80\%$ of queries. It is also noted that metrics with worse performance for higher ranks sometimes give better recalls for lower ranks. For example, the histogram distance with angle information gives higher recalls when ranks lower than 10\% are considered.

\begin{figure}
\centering
\includegraphics[scale=0.4]{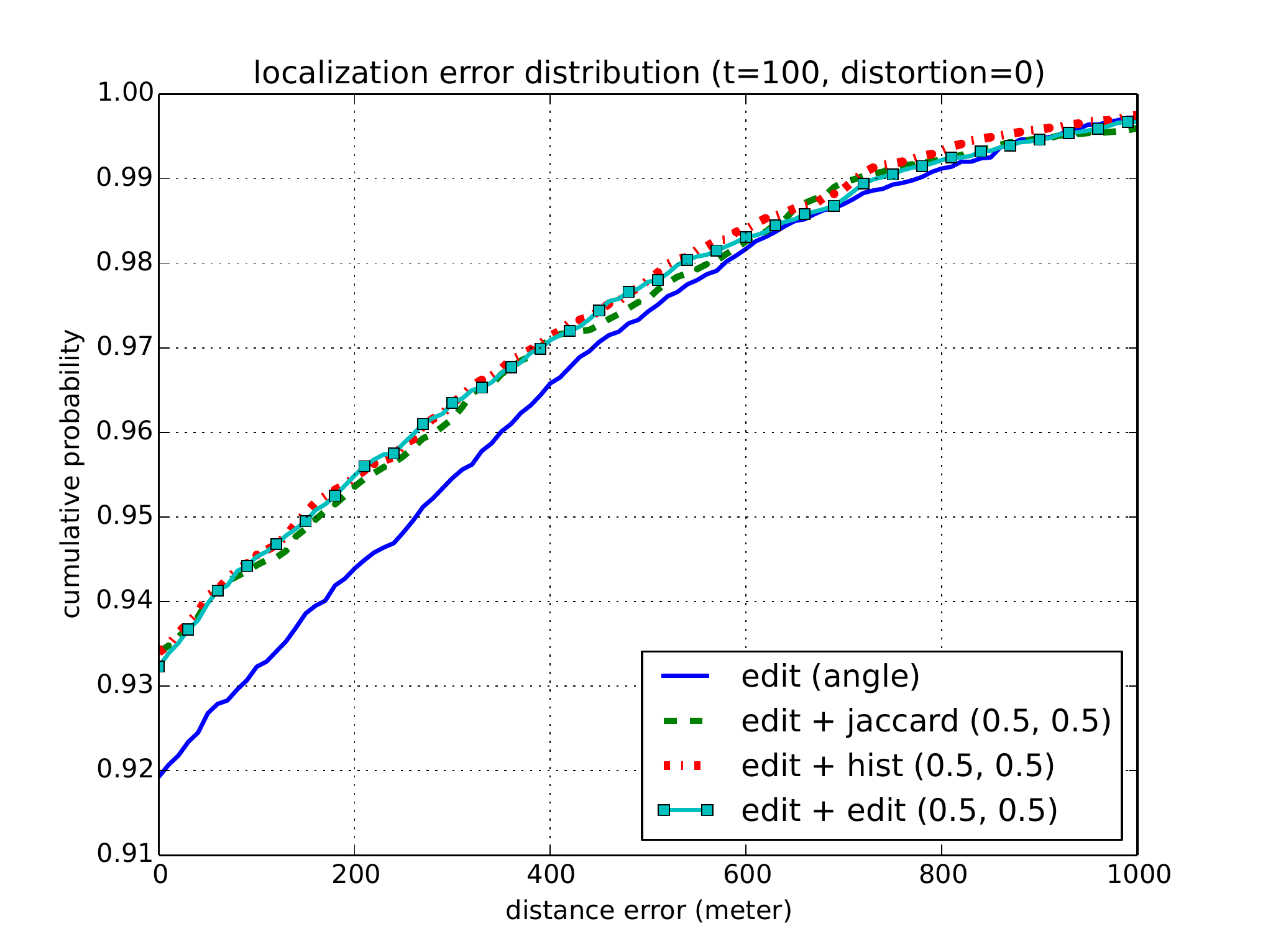}
\caption{Localization error (metric fusion, no distortion).}
\label{fig_localization_metric_fusion_t100_d0}
\end{figure}

\begin{figure}
\centering
\includegraphics[scale=0.4]{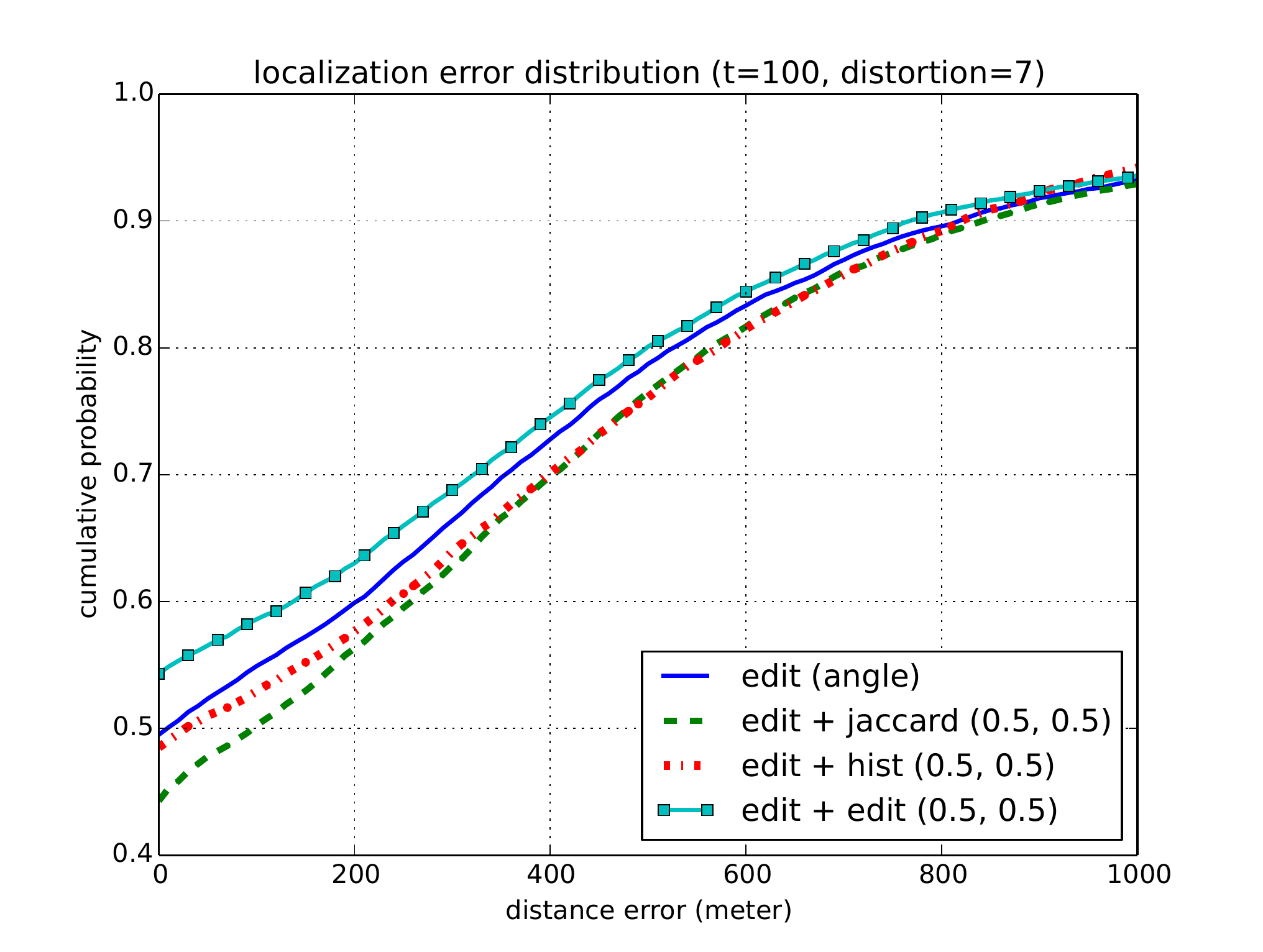}
\caption{Localization error (metric fusion, medium distortion).}
\label{fig_localization_metric_fusion_t100_d7}
\end{figure}

\begin{figure}
\centering
\includegraphics[scale=0.4]{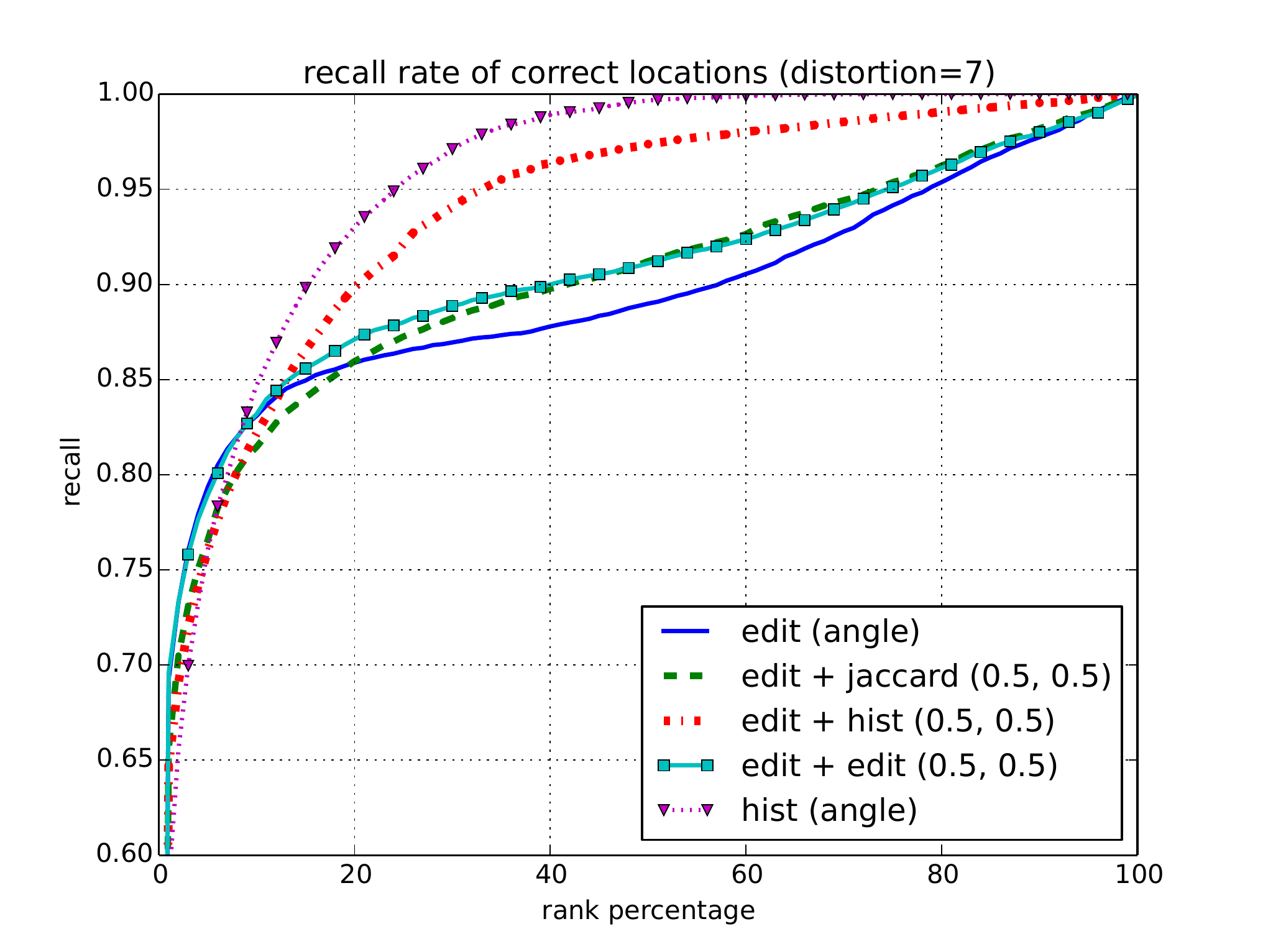}
\caption{Location recall (metric fusion, medium distortion).}
\label{fig_groundtruth_metric_fusion_d7}
\end{figure}

Figure~\ref{fig_localization_metric_fusion_t100_d0} shows the distribution of localization errors for metric fusion. Since edit distance performs best as a single metric, we fix it for type information and try different metrics for angle information. For example, the legend ``edit + jaccard (0.5, 0.5)'' means that edit distance is used for type, and jaccard distance is used for angle; the weight factors are $0.5$ and $0.5$. The figure confirms that combining type and angle information indeed brings performance improvement. The initial probability increases from $0.92$ to more than $0.93$. All combinations seem to perform equally well. When there is distortion (see Fig.~\ref{fig_localization_metric_fusion_t100_d7}), it is more obvious that ``edit + edit'' is the best combination. On the other hand, note that using edit distance with angle information alone even outperforms the other combinations. That means, although angle information is harder to measure, it has stronger discrimination power than type information, which is consistent with the statistics in Table~\ref{tab_signature_database}b. It is also an indication that spatial distribution is useful in localization if properly utilized.
The corresponding recall is shown in Fig.~\ref{fig_groundtruth_metric_fusion_d7}. Compared with Fig.~\ref{fig_groundtruth_single_metric_d7}, the advantage of metric fusion is clear for higher ranks where curves are relatively close; for lower ranks, ``edit + edit'' continues to outperform ``edit (angle)'', but ``edit + hist'' performs worse than ``hist (angle)'' as a trade-off for a slight improvement at higher ranks. We conclude that in general metric fusion is beneficial.

\begin{figure}
\centering
\includegraphics[scale=0.4]{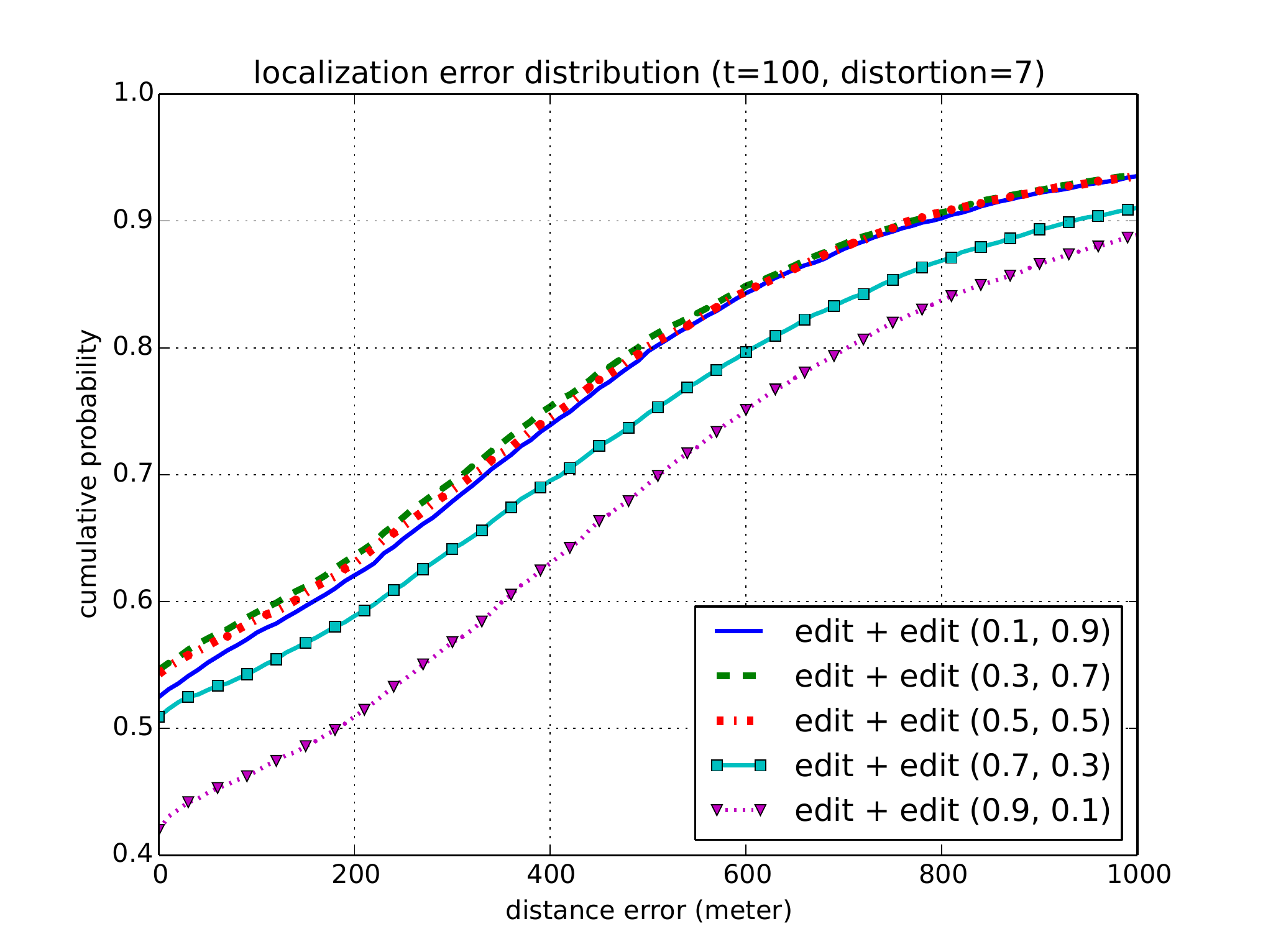}
\caption{Localization error for different weight factors (metric fusion, medium distortion).}
\label{fig_localization_metric_fusion_t100_d7_weight}
\end{figure}

Another important question is what weight factors are the best for metric fusion. In Fig.~\ref{fig_localization_metric_fusion_t100_d7_weight}, various weights are tested. Even weights or slightly larger weights for the angle turn out to be good choices, because settings (0.3, 0.7) and (0.5, 0.5) perform the best. Since type and angle are two independent information sources and the angle performs better when used alone, a slightly higher weight for the angle is reasonable. On the other hand, biasing too much, especially towards the type, such as (0.9, 0.1), decreases the performance. In the following, we keep using even weights (0.5, 0.5).

\begin{figure}
\centering
\includegraphics[scale=0.4]{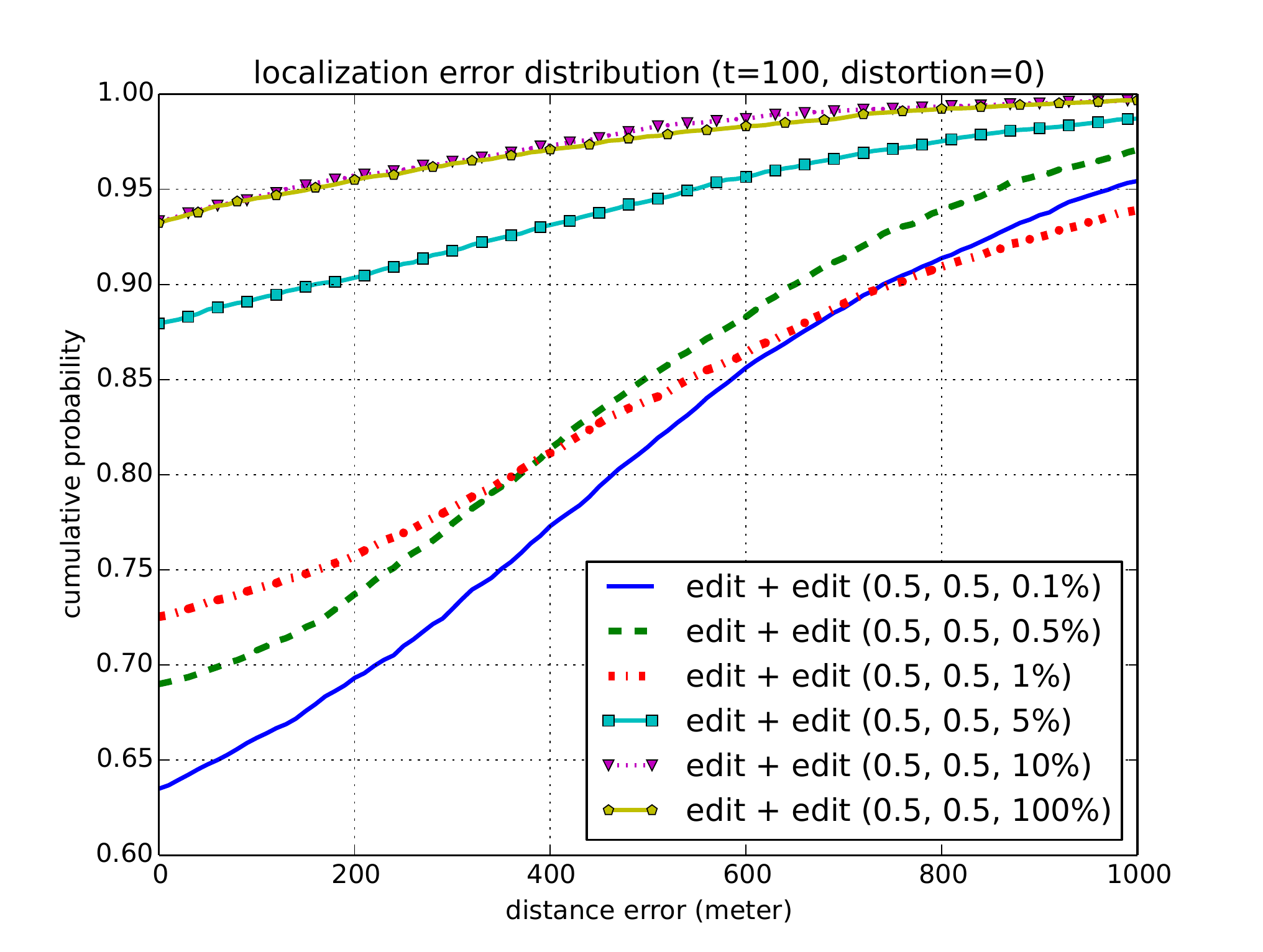}
\caption{Localization error (2-stage metric fusion, no distortion).}
\label{fig_localization_2_stage_metric_fusion_t1_d0}
\end{figure}

\begin{figure}
\centering
\includegraphics[scale=0.4]{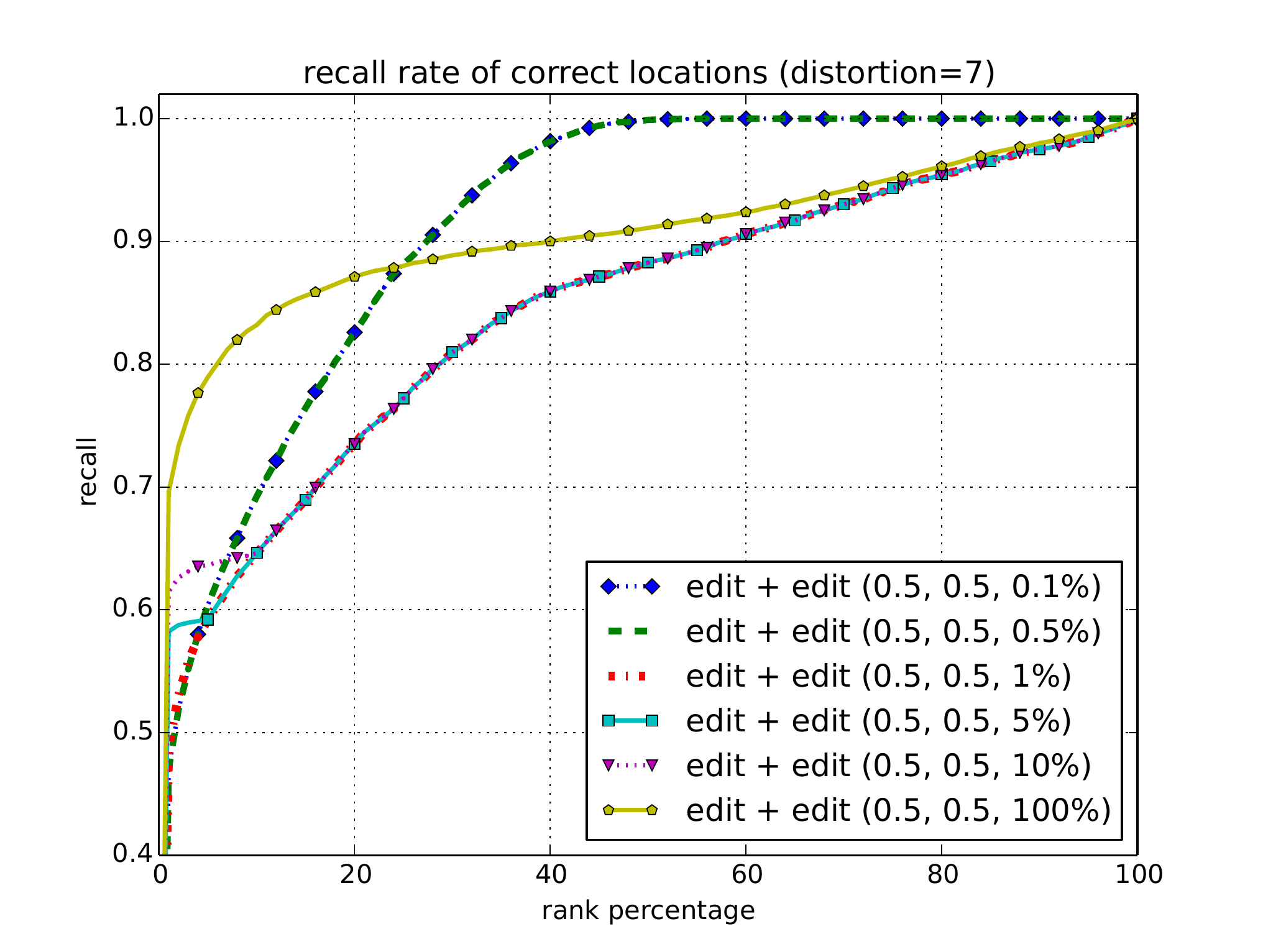}
\caption{Location recall (2-stage metric fusion, medium distortion).}
\label{fig_groundtruth_2_stage_metric_fusion_d7}
\end{figure}

Finally, we look at the results of two-stage metric fusion in Fig.~\ref{fig_localization_2_stage_metric_fusion_t1_d0}. In the legend, the percentage numbers $k\%$ such as $1\%, 5\%$ represent the proportion of candidates for re-ranking. It is interesting that re-ranking less than $10\%$ candidates result in almost the same performance as $100\%$, which implies a significant amount of saving in computation. On the other hand, there is a clear performance drop in recall (see Fig.~\ref{fig_groundtruth_2_stage_metric_fusion_d7}). This is a trade-off between recall and computation efficiency.

\subsubsection{Computation complexity}
We also investigate the computation complexity for different configurations. Table~\ref{tab_retrieval_time} lists the measured retrieval time for a single query with the average length (14). The results are approximate, for only the signature comparison time is counted. They are obtained by averaging over $100$ repetitions using a single metric on a PC with a 3.6 GHz CPU. Clearly, the retrieval time is inversely proportional to the localization performance. For example, edit distance with angle information performs the best among the single metric settings, but it takes the longest time; Jaccard distance does not offer the best localization accuracy, but it is the fastest. For metric fusion, it is straight-forward to estimate the retrieval time by adding up the time for selected metrics. Some example numbers for two-stage metric fusion are listed in the last column of Table~\ref{tab_retrieval_time}. Apparently ``edit + edit'' is the slowest combination, although it provides the best accuracy. Therefore, sometimes it is necessary to make a compromise for speed.
\begin{table}
\centering
\caption{Approximate retrieval time for a single query.}
\begin{tabular}{l|ll|l}
\hline
                  & Type & Angle & Type+Angle, 5\%\\
\hline
Jaccard distance   &  82 ms   &  496 ms & 107 ms\\
Histogram distance &  173 ms  &  538 ms & 200 ms\\
Edit distance      &  277 ms  &  1.85 s & 370 ms\\
\hline
\end{tabular}
\label{tab_retrieval_time}
\end{table}

\subsubsection{Practical localization and parameter dependence}
\begin{figure}
\centering
\includegraphics[scale=0.4]{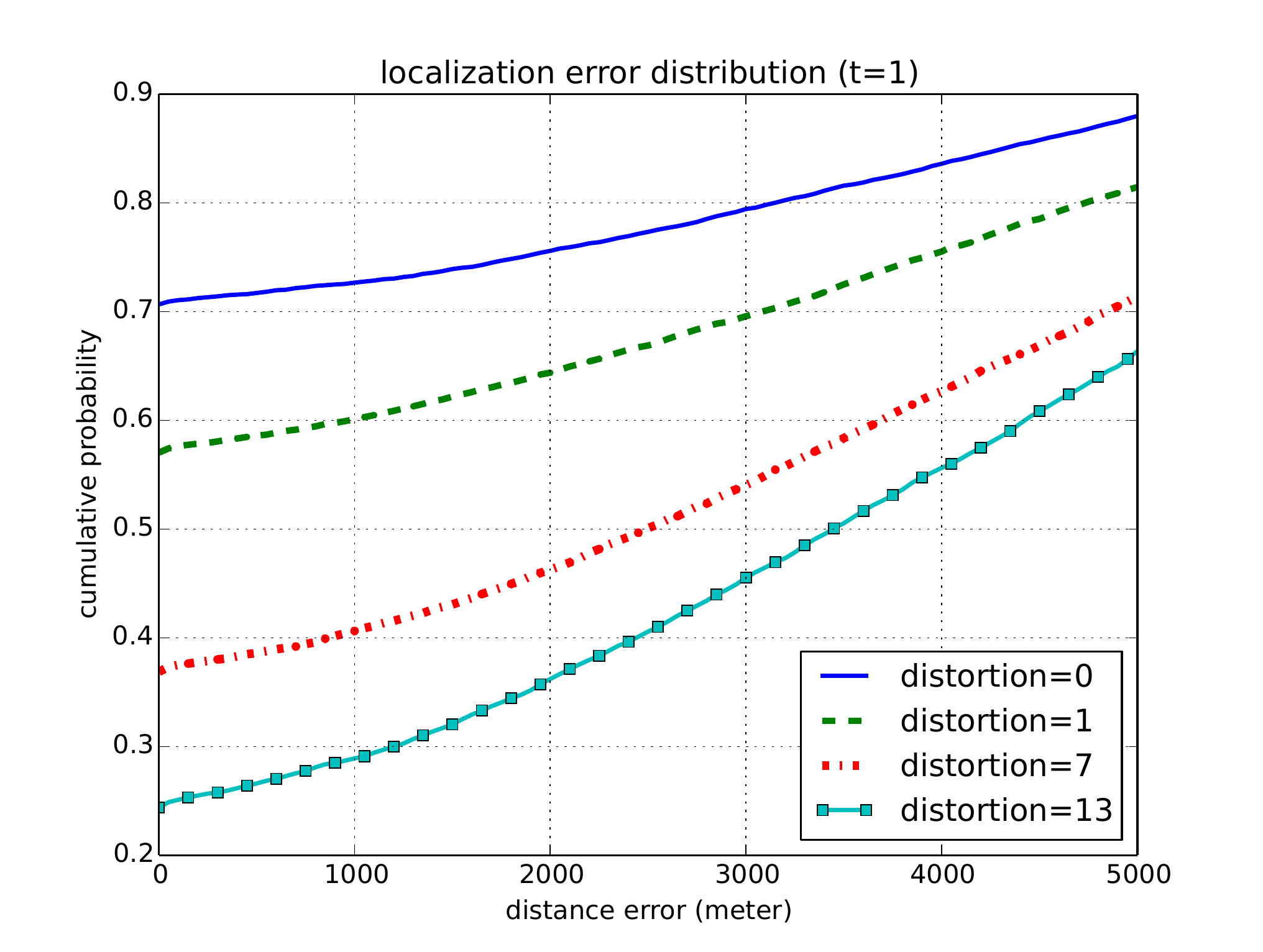}
\caption{Localization error for 2-stage metric fusion ``edit+edit (0.5,0.5) 5\%'', $t=1$.}
\label{fig_localization_2_stage_metric_fusion_t1_d}
\end{figure}

\begin{figure}
\centering
\includegraphics[scale=0.4]{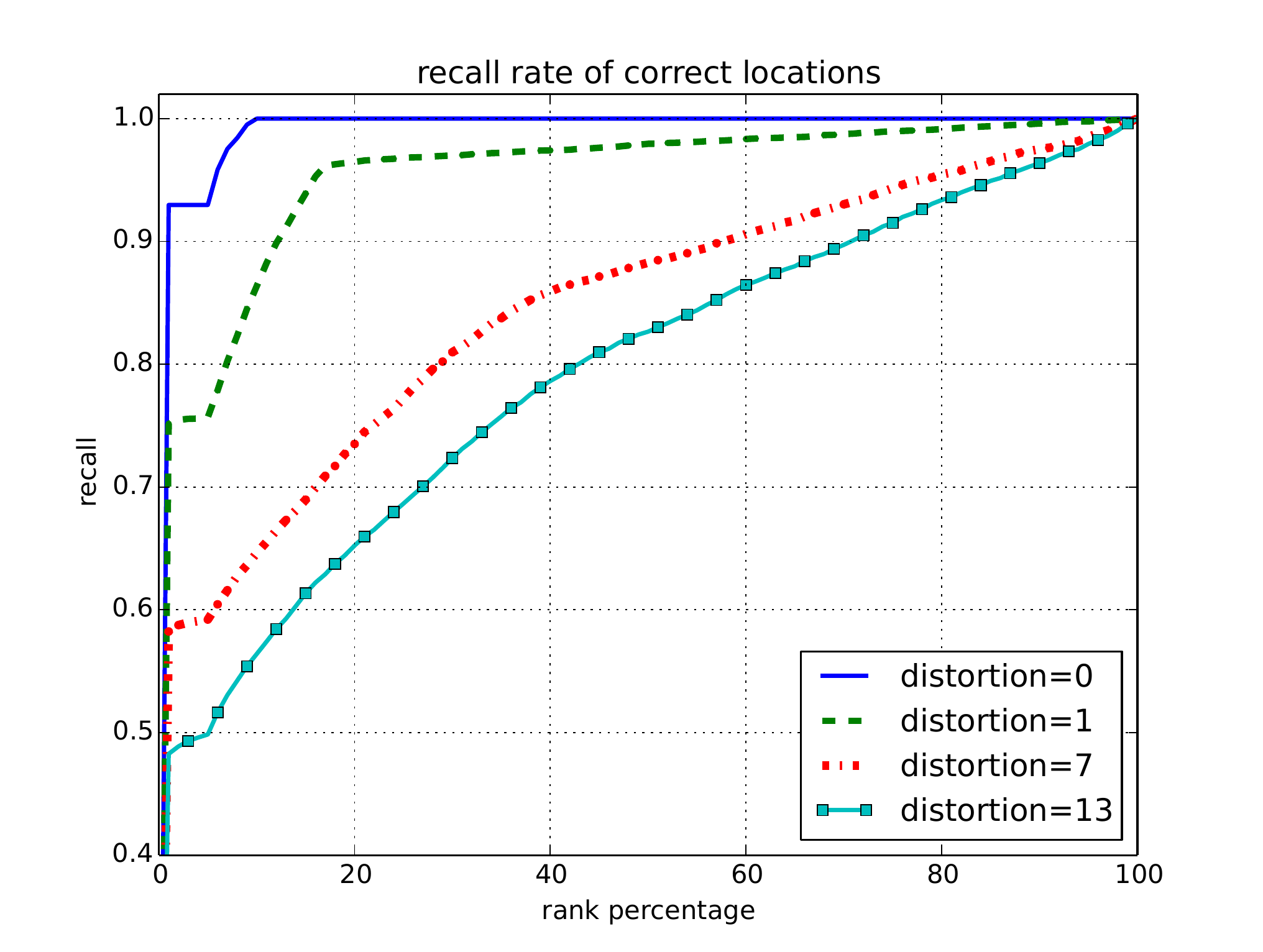}
\caption{Location recall for 2-stage metric fusion ``edit+edit (0.5,0.5) 5\%''.}
\label{fig_groundtruth_2_stage_metric_fusion_d}
\end{figure}

\begin{figure}
\centering
\includegraphics[scale=0.5]{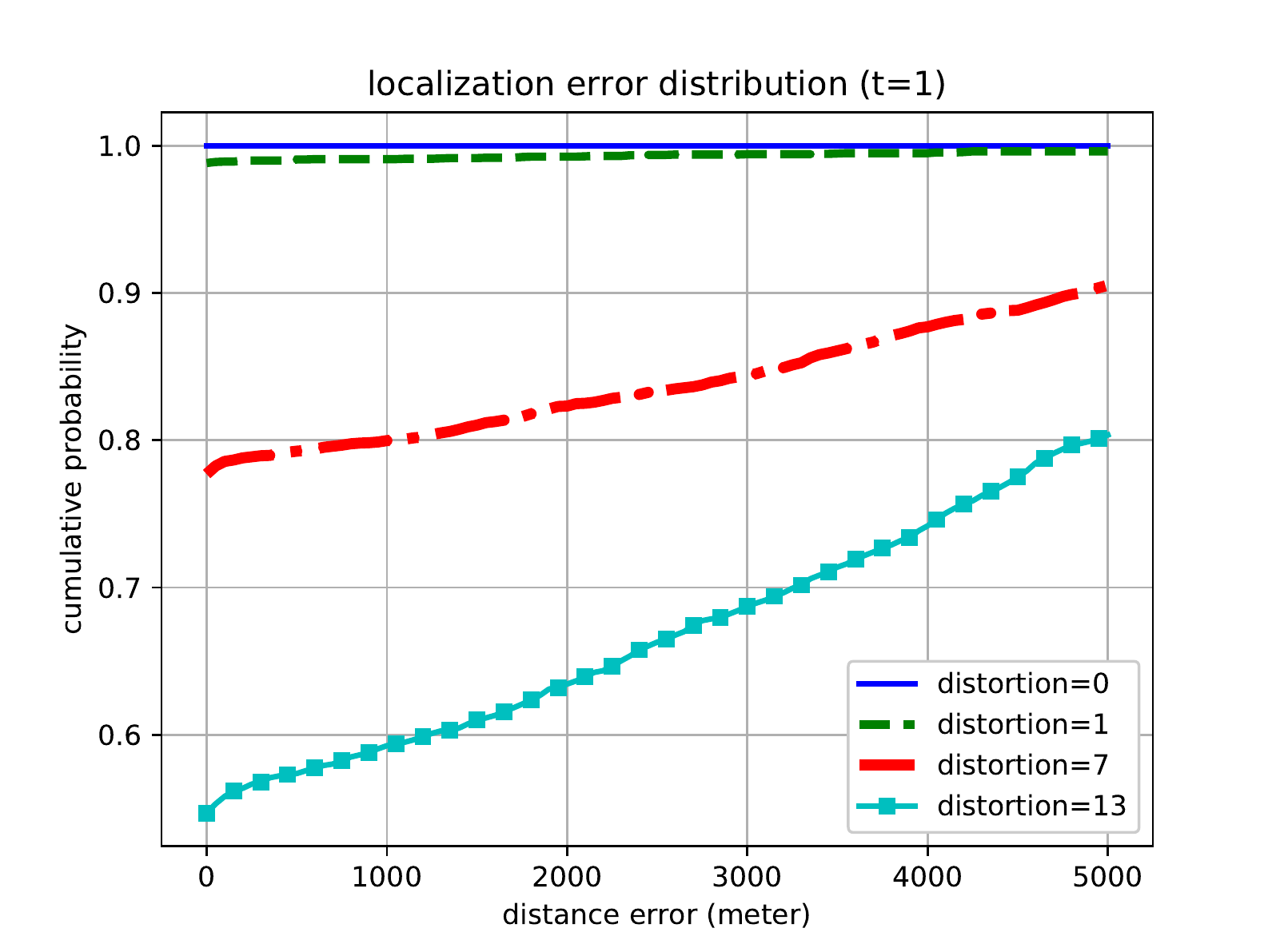}
\caption{Localization error for 2-stage metric fusion ``edit+edit (0.5,0.5) 5\%'', $t=1$. Unambiguous queries are used.}
\label{fig_localization_2_stage_metric_fusion_t1_d_selected}
\end{figure}

\begin{figure}
\centering
\includegraphics[scale=0.5]{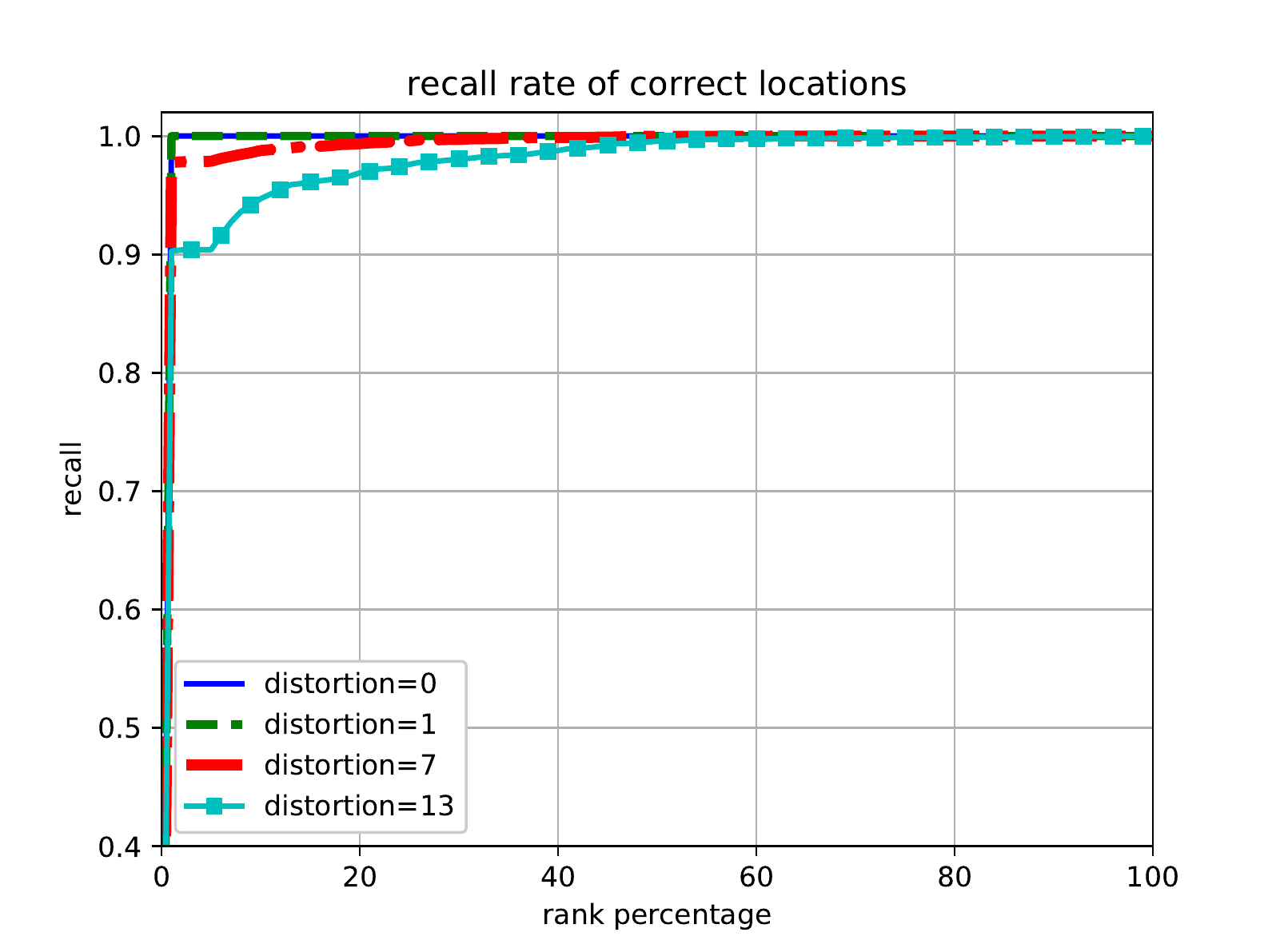}
\caption{Location recall for 2-stage metric fusion ``edit+edit (0.5,0.5) 5\%''. Unambiguous queries are used.}
\label{fig_groundtruth_2_stage_metric_fusion_d_selected}
\end{figure}

Previous results show that two-stage metric fusion achieves a good trade-off between accuracy and speed. Therefore, we consider the configuration ``edit+edit (0.5,0.5) 5\%'' as a practical setting. Figure~\ref{fig_localization_2_stage_metric_fusion_t1_d} shows the localization error distribution for $t=1$ and several distortion levels. This is a ``worst-case'' scenario, because the top candidate location is judged as the query location. The cumulative probability for small distance errors ranges from $0.25$ to $0.70$. This is still an encouraging result. The corresponding recall is shown in Fig.~\ref{fig_groundtruth_2_stage_metric_fusion_d}. The recall for top ranks varies from $0.50$ to more than $0.90$. These results imply the effectiveness of the proposed scheme in practice. 
When used as a filtering tool for other methods such as \cite{Sattler2011,Song2016,Bhowmik2017}, a significant amount of computation and storage might be saved by only considering data associated with top candidate locations. 
In fact, among the ten thousand query signatures, 4180 (41.8\%) have distinct patterns. If only these queries are used, better localization performance could be expected. Additional tests are performed using those ``good'' queries. The results are shown in Fig.~\ref{fig_localization_2_stage_metric_fusion_t1_d_selected}--\ref{fig_groundtruth_2_stage_metric_fusion_d_selected}. They are significantly improved compared with previous ones, which again confirms the value of our scheme.

We further examine the performance dependence on some system parameters with the same two-stage metric fusion setting ``edit+edit (0.5,0.5) 5\%, t=1''. In particular, we focus on the visibility range and the angle quantization level. They both depend on the camera and the object detection algorithm. The visibility range is generally proportional to the signature length -- the larger visibility, the longer signatures. As the signature entropy increases, it should be easier to distinguish one location from another. This is confirmed by the results in Table~\ref{tab_performance_dependence_visibility}, where the cumulative probability and the recall is shown for different visibility ranges, with a fixed maximum localization error (50m) and a fixed rank range (top 10\%). The results are obtained by generating new databases and repeating the experiments. They suggest that localization performance can be improved by using more powerful imaging devices and algorithms.

On the other hand, angle quantization is meant to counteract noise, e.g. measurement errors. The stronger quantization, the more resistance to noise (and potentially less storage and computation), but also less discrimination power. Table~\ref{tab_performance_dependence_quantization} shows the cumulative probability for localization errors up to 50m with various quantization strengths. The performance first increases with the number of quantization levels, then slightly drops, which indicates that finer quantization does not always bring better performance. Thus a balance should be sought between noise resistance and discrimination power.

\begin{table}
\centering
\caption{Performance dependence on visibility. $R$ denotes the visibility range.}
\begin{tabular}{l|l|l}
\hline
          &  P(error$\leq$50m) & recall@10\% \\
\hline
$R=20$    & $0.1793$ & $0.5110$\\
$R=30$    & $0.3733$ & $0.6462$\\
$R=40$    & $0.4946$ & $0.7269$\\
\hline
\end{tabular}
\label{tab_performance_dependence_visibility}
\end{table}

\begin{table}
\centering
\caption{Performance dependence on angle quantization. $Q$ denotes the number of quantization levels.}
\begin{tabular}{l|l}
\hline
          & P(error$\leq$50m) \\
\hline
$Q=8$ & $0.3243$  \\
$Q=16$ & $0.3733$  \\
$Q=24$ & $0.3861$  \\
$Q=32$ & $0.3766$  \\
\hline
\end{tabular}
\label{tab_performance_dependence_quantization}
\end{table}

\section{Conclusion and discussion}
\label{sect_conclusion}
In this work, we propose to use semantic information for urban localization. We focus on special objects that can be seen from street views, such as trees, street lights, bus stops, etc. These semantic objects can be obtained from public data sources. They are encoded as semantic signatures. The localization problem is solved by signature matching. Given a query signature, similar signatures are retrieved from a database. The query location is inferred from the best matches' geo-reference. A semantic signature consists of two parts, a type sequence and an angle sequence. We select a few metrics for sequence matching and find that edit distance shows promising results. In order to aggregate both type and angle information, a metric fusion framework is proposed for signature retrieval. In addition, a two-stage fusion approach is proposed to improve computation efficiency.

Simulation shows that the proposed technique ideally achieves close-to-GPS accuracy. In practice, it can be used alone for coarse localization, and also in integration with other techniques for more accurate localization, such as pose estimation. It is interesting for e.g. tourism applications in urban areas. Since the scheme can effectively filter out irrelevant regions, it is a suitable step before other matching techniques that require heavy computation.

This paper focuses on retrieval. A number of existing semantic objects are used. While object detection is not covered here, the main message of this self-contained work is that if a sufficient amount of semantic objects exist, then satisfactory localization is possible even in a large scale. There are also other obstacles in reality, such as inaccurate distance measurement, object occlusion, out-of-date databases etc. These issues are partly taken into account by simulated distortion. Fine-tuning the system together with an object detection pipeline is an interesting topic for future research.


\bibliographystyle{IEEEtran}      
\bibliography{semantic_signature_localization}   

\begin{thebibliography}{10}
\providecommand{\url}[1]{#1}
\csname url@samestyle\endcsname
\providecommand{\newblock}{\relax}
\providecommand{\bibinfo}[2]{#2}
\providecommand{\BIBentrySTDinterwordspacing}{\spaceskip=0pt\relax}
\providecommand{\BIBentryALTinterwordstretchfactor}{4}
\providecommand{\BIBentryALTinterwordspacing}{\spaceskip=\fontdimen2\font plus
\BIBentryALTinterwordstretchfactor\fontdimen3\font minus
  \fontdimen4\font\relax}
\providecommand{\BIBforeignlanguage}[2]{{%
\expandafter\ifx\csname l@#1\endcsname\relax
\typeout{** WARNING: IEEEtran.bst: No hyphenation pattern has been}%
\typeout{** loaded for the language `#1'. Using the pattern for}%
\typeout{** the default language instead.}%
\else
\language=\csname l@#1\endcsname
\fi
#2}}
\providecommand{\BIBdecl}{\relax}
\BIBdecl

\bibitem{Lowry2016}
S.~Lowry, N.~S{\"u}nderhauf, P.~Newman, J.~J. Leonard, D.~Cox, P.~Corke, and
  M.~J. Milford, ``Visual place recognition: A survey,'' \emph{IEEE
  Transactions on Robotics}, vol.~32, no.~1, pp. 1--19, Feb. 2016.

\bibitem{Piasco2018}
N.~Piasco, D.~Sidib{\'e}, C.~Demonceaux, and V.~Gouet-Brunet, ``A survey on
  visual-based localization: On the benefit of heterogeneous data,''
  \emph{Pattern Recognition}, vol.~74, pp. 90--109, 2018.

\bibitem{Lim2012}
H.~Lim, S.~N. Sinha, M.~F. Cohen, and M.~Uyttendaele, ``Real-time image-based
  6-{DOF} localization in large-scale environments,'' in \emph{Proc. of IEEE
  Conference on Computer Vision and Pattern Recognition (CVPR)}, June 2012, pp.
  1043--1050.

\bibitem{Snavely2006}
N.~Snavely, S.~M. Seitz, and R.~Szeliski, ``Photo tourism: Exploring photo
  collections in {3D},'' \emph{ACM Trans. Graph.}, vol.~25, no.~3, pp.
  835--846, Jul. 2006.

\bibitem{Crandall2009}
D.~J. Crandall, L.~Backstrom, D.~Huttenlocher, and J.~Kleinberg, ``Mapping the
  world's photos,'' in \emph{Proc. of International Conference on World Wide
  Web (WWW)}.\hskip 1em plus 0.5em minus 0.4em\relax ACM, 2009, pp. 761--770.

\bibitem{Qu2015}
X.~Qu, B.~Soheilian, and N.~Paparoditis, ``Vehicle localization using
  mono-camera and geo-referenced traffic signs,'' in \emph{Proc. of IEEE
  Intelligent Vehicles Symposium}, June 2015, pp. 605--610.

\bibitem{Brachmann2018}
E.~Brachmann and C.~Rother, ``Learning less is more - 6d camera localization
  via 3d surface regression,'' in \emph{IEEE Conference on Computer Vision and
  Pattern Recognition (CVPR)}, 2018.

\bibitem{Lowe2004}
D.~G. Lowe, ``Distinctive image features from scale-invariant keypoints,''
  \emph{International Journal on Computer Vision (IJCV)}, vol.~60, no.~2, pp.
  91--110, Nov. 2004.

\bibitem{Schindler2007}
G.~Schindler, M.~Brown, and R.~Szeliski, ``City-scale location recognition,''
  in \emph{Proc. of IEEE Conference on Computer Vision and Pattern Recognition
  (CVPR)}, June 2007, pp. 1--7.

\bibitem{Li2010}
Y.~Li, N.~Snavely, and D.~P. Huttenlocher, ``Location recognition using
  prioritized feature matching,'' in \emph{Proc. of European Conference on
  Computer Vision (ECCV)}, 2010, pp. 791--804.

\bibitem{Sattler2011}
T.~Sattler, B.~Leibe, and L.~Kobbelt, ``Fast image-based localization using
  direct {2D-to-3D} matching,'' in \emph{Proc. of International Conference on
  Computer Vision (ICCV)}, Nov 2011, pp. 667--674.

\bibitem{Li2012}
Y.~Li, N.~Snavely, D.~Huttenlocher, and P.~Fua, ``Worldwide pose estimation
  using {3D} point clouds,'' in \emph{Proc. of European Conference on Computer
  Vision (ECCV)}, 2012, pp. 15--29.

\bibitem{Lin2017}
T.~{Lin}, P.~{Goyal}, R.~{Girshick}, K.~{He}, and P.~{Dollár}, ``Focal loss
  for dense object detection,'' in \emph{2017 IEEE International Conference on
  Computer Vision (ICCV)}, Oct 2017, pp. 2999--3007.

\bibitem{Redmon2018}
\BIBentryALTinterwordspacing
J.~Redmon and A.~Farhadi, ``Yolov3: An incremental improvement,'' \emph{CoRR},
  vol. abs/1804.02767, 2018. [Online]. Available:
  \url{http://arxiv.org/abs/1804.02767}
\BIBentrySTDinterwordspacing

\bibitem{Weng2018}
L.~{Weng}, B.~{Soheilian}, and V.~{Gouet-Brunet}, ``Semantic signatures for
  urban visual localization,'' in \emph{International Conference on
  Content-Based Multimedia Indexing (CBMI)}, Sep. 2018, pp. 1--6.

\bibitem{Nister2006}
D.~Nister and H.~Stewenius, ``Scalable recognition with a vocabulary tree,'' in
  \emph{Proc. of IEEE Conference on Computer Vision and Pattern Recognition
  (CVPR)}, vol.~2, 2006, pp. 2161--2168.

\bibitem{Irschara2009}
A.~Irschara, C.~Zach, J.~M. Frahm, and H.~Bischof, ``From structure-from-motion
  point clouds to fast location recognition,'' in \emph{Proc. of IEEE
  Conference on Computer Vision and Pattern Recognition (CVPR)}, June 2009, pp.
  2599--2606.

\bibitem{Arya1993}
S.~Arya and D.~M. Mount, ``Approximate nearest neighbor queries in fixed
  dimensions,'' in \emph{Proc. of ACM-SIAM Symposium on Discrete Algorithms
  (SODA)}, 1993, pp. 271--280.

\bibitem{Zamir2010}
A.~R. Zamir and M.~Shah, ``Accurate image localization based on google maps
  street view,'' in \emph{Proc. of European Conference on Computer Vision
  (ECCV)}, 2010, pp. 255--268.

\bibitem{Chen2011}
D.~M. Chen, G.~Baatz, K.~K\"oser, S.~S. Tsai, R.~Vedantham, T.~Pylv\"an\"ainen,
  K.~Roimela, X.~Chen, J.~Bach, M.~Pollefeys, B.~Girod, and R.~Grzeszczuk,
  ``City-scale landmark identification on mobile devices,'' in \emph{Proc. of
  IEEE Conference on Computer Vision and Pattern Recognition (CVPR)}, June
  2011, pp. 737--744.

\bibitem{Zhang2011}
J.~Zhang, A.~Hallquist, E.~Liang, and A.~Zakhor, ``Location-based image
  retrieval for urban environments,'' in \emph{Proc. of IEEE International
  Conference on Image Processing (ICIP)}, 2011, pp. 3677--3680.

\bibitem{Fischler1981}
M.~A. Fischler and R.~C. Bolles, ``Random sample consensus: A paradigm for
  model fitting with applications to image analysis and automated
  cartography,'' \emph{Commun. ACM}, vol.~24, no.~6, pp. 381--395, Jun. 1981.

\bibitem{Tola2008}
E.~Tola, V.~Lepetit, and P.~Fua, ``A fast local descriptor for dense
  matching,'' in \emph{Proc. of IEEE Conference on Computer Vision and Pattern
  Recognition (CVPR)}, June 2008, pp. 1--8.

\bibitem{Calonder2010}
M.~Calonder, V.~Lepetit, C.~Strecha, and P.~Fua, ``Brief: Binary robust
  independent elementary features,'' in \emph{ECCV}, 2010, pp. 778--792.

\bibitem{Shrivastava2011}
A.~Shrivastava, T.~Malisiewicz, A.~Gupta, and A.~A. Efros, ``Data-driven visual
  similarity for cross-domain image matching,'' \emph{ACM Trans. Graph.},
  vol.~30, no.~6, p.~10, Dec. 2011.

\bibitem{Zamir2014}
A.~R. Zamir and M.~Shah, ``Image geo-localization based on multiple nearest
  neighbor feature matching using generalized graphs,'' \emph{IEEE Transactions
  on Pattern Analysis and Machine Intelligence}, vol.~36, no.~8, pp.
  1546--1558, Aug 2014.

\bibitem{Arandjelovic2014}
R.~Arandjelovi{\'{c}} and A.~Zisserman, ``Dislocation: Scalable descriptor
  distinctiveness for location recognition,'' in \emph{Proc. of Asian
  Conference on Computer Vision (ACCV)}, 2014, pp. 188--204.

\bibitem{Torii2015}
A.~Torii, R.~Arandjelovic, J.~Sivic, M.~Okutomi, and T.~Pajdla, ``24/7 place
  recognition by view synthesis,'' in \emph{Proc. of IEEE Conference on
  Computer Vision and Pattern Recognition (CVPR)}, 2015, pp. 1808--1817.

\bibitem{Jegou2010}
H.~J\'egou, M.~Douze, C.~Schmid, and P.~P\'erez, ``Aggregating local
  descriptors into a compact image representation,'' in \emph{Proc. of IEEE
  Conference on Computer Vision and Pattern Recognition (CVPR)}, June 2010, pp.
  3304--3311.

\bibitem{Arandjelovic2016}
R.~Arandjelovic, P.~Gronat, A.~Torii, T.~Pajdla, and J.~Sivic, ``{NetVLAD: CNN}
  architecture for weakly supervised place recognition,'' in \emph{Proc. of
  IEEE Conference on Computer Vision and Pattern Recognition (CVPR)}, 2016, pp.
  5297--5307.

\bibitem{Iscen2017}
A.~Iscen, G.~Tolias, Y.~Avrithis, T.~Furon, and O.~Chum, ``Panorama to panorama
  matching for location recognition,'' in \emph{Proc. of ACM International
  Conference on Multimedia Retrieval}, 2017, pp. 392--396.

\bibitem{Agarwal2009}
S.~Agarwal, N.~Snavely, I.~Simon, S.~M. Seitz, and R.~Szeliski, ``Building
  {Rome} in a day,'' in \emph{Proc. of International Conference on Computer
  Vision (ICCV)}, Sept 2009, pp. 72--79.

\bibitem{Song2016}
Y.~Song, X.~Chen, X.~Wang, Y.~Zhang, and J.~Li, ``6-{DOF} image localization
  from massive geo-tagged reference images,'' \emph{IEEE Transactions on
  Multimedia}, vol.~18, no.~8, pp. 1542--1554, Aug 2016.

\bibitem{Sattler2017}
T.~Sattler, A.~Torii, J.~Sivic, M.~Pollefeys, H.~Taira, M.~Okutomi, and
  T.~Pajdla, ``Are large-scale 3d models really necessary for accurate visual
  localization?'' in \emph{Proc. of IEEE Conference on Computer Vision and
  Pattern Recognition (CVPR)}, 2017, p.~10.

\bibitem{Ardeshir2014}
S.~Ardeshir, A.~R. Zamir, A.~Torroella, and M.~Shah, ``{GIS}-assisted object
  detection and geospatial localization,'' in \emph{Proc. of Eupropean
  Conference on Computer Vision (ECCV)}, 2014, pp. 602--617.

\bibitem{Arth2015}
C.~Arth, C.~Pirchheim, J.~Ventura, D.~Schmalstieg, and V.~Lepetit, ``Instant
  outdoor localization and slam initialization from {2.5D} maps,'' in
  \emph{Proc. of International Symposium on Mixed and Augmented Reality
  (ISMAR)}, 2015.

\bibitem{Bhowmik2017}
N.~Bhowmik, L.~Weng, V.~Gouet-Brunet, and B.~Soheilian, ``Cross-domain image
  localization by adaptive feature fusion,'' in \emph{Proc. of Joint Urban
  Remote Sensing Event}, 2017, p.~4.

\bibitem{Jaccard1912}
P.~Jaccard, ``The distribution of the flora in the alpine zone,'' \emph{New
  Phytologist}, vol.~11, no.~2, pp. 37--50, 1912.

\bibitem{Navarro2001}
G.~Navarro, ``A guided tour to approximate string matching,'' \emph{ACM
  Computing Surveys}, vol.~33, no.~1, pp. 31--88, 2001.

\bibitem{Snoek2005}
\BIBentryALTinterwordspacing
C.~G.~M. Snoek, M.~Worring, and A.~W.~M. Smeulders, ``Early versus late fusion
  in semantic video analysis,'' in \emph{ACM International Conference on
  Multimedia}, New York, NY, USA, 2005, pp. 399--402. [Online]. Available:
  \url{http://doi.acm.org/10.1145/1101149.1101236}
\BIBentrySTDinterwordspacing

\bibitem{Vrochidis2019}
S.~Vrochidis, B.~Huet, E.~Y. Chang, and I.~Kompatsiaris, Eds., \emph{Big Data
  Analytics for Large-Scale Multimedia Search}.\hskip 1em plus 0.5em minus
  0.4em\relax Wiley, 2019.

\bibitem{Girshick2015}
R.~Girshick, ``Fast {R-CNN},'' in \emph{Proc. of IEEE International Conference
  on Computer Vision and Pattern Recognition}, 2015, pp. 1440--1448.

\bibitem{He2015}
K.~He, X.~Zhang, S.~Ren, and J.~Sun, ``Spatial pyramid pooling in deep
  convolutional networks for visual recognition,'' \emph{IEEE Transactions on
  Pattern Analysis and Machine Intelligence}, vol.~37, no.~9, pp. 1904--1916,
  2015.

\bibitem{Ren2015}
S.~Ren, K.~He, R.~Girshick, and J.~Sun, ``Faster {R-CNN}: Towards real-time
  object detection with region proposal networks,'' in \emph{Advances in neural
  information processing systems (NIPS)}, 2015, pp. 91--99.

\end{thebibliography}







\end{document}